\documentclass[10pt,twocolumn,letterpaper]{article}

\usepackage[pagenumbers]{cvpr} 

\usepackage{algorithm}
\usepackage{algorithmic}

\usepackage{graphicx}
\usepackage{amsmath}
\usepackage{amssymb}
\usepackage{booktabs}
\usepackage{graphicx}
\usepackage{multirow}
\usepackage{makecell}
\usepackage{arydshln}
\usepackage[export]{adjustbox}
\usepackage[accsupp]{axessibility}  

\usepackage{pifont}
\usepackage{xcolor}

\usepackage[pagebackref,breaklinks,colorlinks]{hyperref}
\newcommand{\tabincell}[2]{\begin{tabular}{@{}#1@{}}#2\end{tabular}}

\usepackage[capitalize]{cleveref}
\crefname{section}{Sec.}{Secs.}
\Crefname{section}{Section}{Sections}
\Crefname{table}{Table}{Tables}
\crefname{table}{Tab.}{Tabs.}

\def\cvprPaperID{6376} 
\def\confName{CVPR}
\def\confYear{2023}

\begin{document}

\title{SwinFIR: Revisiting the SwinIR with Fast Fourier Convolution and Improved Training for Image Super-Resolution}

\author{ Dafeng Zhang$^{1}$\thanks{Equal contribution.}\hspace{20pt}  Feiyu Huang$^1$\footnotemark[1]  \hspace{20pt} Shizhuo Liu$^1$\hspace{20pt} Xiaobing Wang$^1$\hspace{20pt} Zhezhu Jin$^1$\\ 
{$^1 $} Samsung Research China - Beijing (SRC-B) \\
{\tt\small \{dfeng.zhang,feiyu.huang, shizhuo.liu, x0106.wang, zz777.jin\}@samsung.com}
}

\maketitle

\begin{abstract}
Transformer-based methods have achieved impressive image restoration performance due to their capacities to model long-range dependency compared to CNN-based methods. However, advances like SwinIR adopts the window-based and local attention strategy to balance the performance and computational overhead, which restricts employing large receptive fields to capture global information and establish long dependencies in the early layers. To further improve the efficiency of capturing global information, in this work, we propose SwinFIR to extend SwinIR by replacing Fast Fourier Convolution (FFC) components, which have the image-wide receptive field. We also revisit other advanced techniques, \ie, data augmentation, pre-training, and feature ensemble to improve the effect of image reconstruction. And our feature ensemble method enables the performance of the model to be considerably enhanced without increasing the training and testing time. We applied our algorithm on multiple popular large-scale benchmarks and achieved state-of-the-art performance comparing to the existing methods. For example, our SwinFIR achieves the PSNR of 32.83 dB on Manga109 dataset, which is 0.8 dB higher than the state-of-the-art SwinIR method, a significant improvement.
\end{abstract}

\section{Introduction}
Deep learning has been increasingly used for the image super-resolution in recent years. And the performance has significantly improved as a result of increasing network depth~\cite{lim2017enhanced}, recursive learning using ResBlock~\cite{kim2016deeply} and channel attention~\cite{zhang2018image}. Due to the limitations of the receptive field, Convolutional Neural Network (CNN) concentrates on the limited area of the image. Conversely, the attention module can more effectively combine global information in the early layers, which is why it achieves better performance than CNN. Therefore, the network structure based on self-attention, especially Transformer ~\cite{vaswani2017attention, liu2021swin}, can effectively utilize global information from shallow layers to deep.

\begin{figure}[t]
	\centering
	\includegraphics[width=1.0\linewidth]{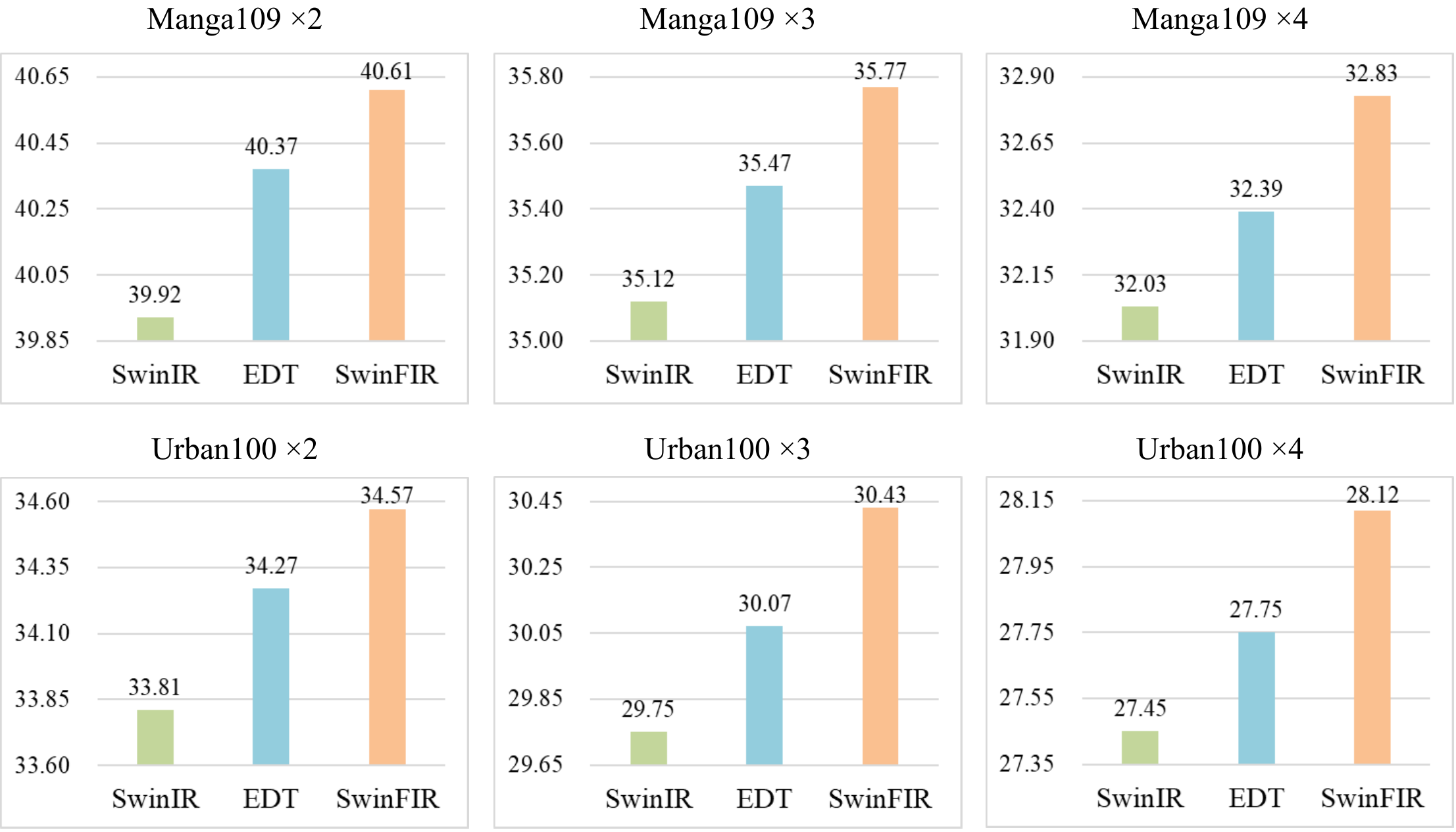}
	\vspace{-6mm}
	\caption{The comparison results of our SwinFIR with the state-of-the-art methods SwinIR and EDT. Under different scales ($\times 2, \times 3, \times 4$), our SwinFIR achieves the best performance than existing works.} 
	\label{fig:histogram}
\end{figure}

The Vision Transformer (VIT)~\cite{dosovitskiy2020image, carion2020end, wang2021pyramid, guo2021sotr, liu2021swin}  has achieved great success in the high-level vision task. Therefore, Liang~\etal~\cite{liang2021swinir} explore the potential of VIT in the low-level vision tasks and propose SwinIR. SwinIR, which is based on Swin Transformer~\cite{liu2021swin}, outperforms the state-of-the-art methods on image restoration tasks such as image super-resolution, and image denoising. It consists of three components: shallow feature extraction, deep feature extraction and high-quality image reconstruction. The fundamental unit of deep feature extraction is called RSTB (Residual Swin Transformer Block), and each RSTB is made up of many Swin Transformer layers and residual connections. In detail, the RSTB performs the self-attention in each window and uses Shift Window strategy to expand the receptive field.  However, this limited shift cannot effectively perceive the global information in the early layers.

Global information is essential for image super-resolution (SR) since it can activate more pixels and is beneficial to improve the image reconstruction performance~\cite{gu2021interpreting}. Therefore, in order to utilize global information, we revisit the SwinIR architecture and introduce a new model specifically designed for SR task, called SwinFIR. The Spatial Frequency Block (SFB), which is based on Fast Fourier Convolution (FFC)~\cite{chi2020fast} and substitutes the convolution layer of the deep feature extraction module of SwinIR, is the essential innovation for SwinFIR. SFB consists of two branches: spatial and frequency model. We employ the FFC to extract the global information in the frequency branch and the CNN-based residual module in the spatial branch to enhance local feature expression.

In addition to the SFB module, we also revisit a variety of methods to improve the image super-resolution performance, such as data augmentation, loss function, pre-training strategy, post-processing, \etc. Data Augmentation (DA) based on the pixel-domain, which is extensively used and has yielded impressive results in high-level tasks, is rarely studied in SR (super-resolution) tasks.  A lot of works~\cite{devries2017improved, hendrycks2019benchmarking, yun2019cutmix, zhang2017mixup} have proved that effective DA can inhibit overfitting and improve the generalization ability of the model. Therefore, we believe that exploring effective DA will certainly boost the effectiveness of image super-resolution. And we demonstrate through experiments that efficient data augmentation approaches, such as channel shuffle and Mixup, can considerably enhance the performance of image super-resolution. At the same time, we propose a brand-new feature-ensemble post-processing technique that is inspired by self-ensemble. The feature ensemble method enables the performance of the model to be considerably enhanced without increasing the training and testing time. 

The comparison results of our SwinFIR with the state-of-the-art methods SwinIR~\cite{ liang2021swinir} and EDT~\cite{li2021efficient} on Manga109 and Urban100 datasets as shown in Figure~\ref{fig:histogram}. Experimental results demonstrate that these strategies can effectively improve the performance of image super-resolution and our SwinFIR achieves state-of-the-art (SOTA) performance on all benchmarks. Specifically, our SwinFIR is 0.30 $\sim$ 0.80 dB and 0.24 $\sim$ 0.44 dB higher than the SOTA methods of SwinIR and EDT on the Manga109 and the Urban100 dataset, respectively, by using these strategies.

Our contributions can be summarized as follows:
\begin{itemize}
\item We revisit the SwinIR architecture and introduce the Spatial Frequency Block (SFB) specifically designed for utilizing global information in SR task, called SwinFIR. SFB is based on Fast Fourier Convolution (FFC) and used extract more comprehensive, detailed, and stable features. SFB consists of two branches: spatial and frequency model. We employ the FFC to extract the global information in the frequency branch and the residual module in the spatial branch to enhance local feature expression.

\item We revisit various data augmentation methods in low-level tasks and demonstrate that efficient data augmentation approaches, such as channel shuffle and mixup, can considerably boost the performance of image super-resolution. Our method breaks the inertial thinking that data enhancement methods such as inserting new pixels will affect SR performance.

\item We propose a brand-new ensemble strategy called feature ensemble, which integrates multiple trained models to get a better and more comprehensive model without increasing training and testing time, and is a zero-cost method to improve performance.
\end{itemize}

\section{Related Works}
Image Super Resolution (SR) is defined as the process of restoring a High Resolution (HR) image from a Low Resolution (LR) image. In recent years, SR models have been actively explored and achieved state-of-the-art performance with the rapid development of deep learning technology. SRCNN~\cite{dong2015image} is the pioneering work of deep learning in SR. The network structure is very simple and only three convolutional layers are used. 
EDSR~\cite{lim2017enhanced} improves performance by removing unnecessary modules in residual networks and expanding the model size. RCAN~\cite{zhang2018image} proposes the Channel Attention mechanism to adaptively rescale features of each channel by modeling the interdependencies between feature channels. 
HAN~\cite{niu2020single} further explores the application of attention mechanisms in SR task by modeling the interdependencies between different layers, different channels and different locations. 
Although all of these CNN-based works achieves excellent performance in SR tasks, CNN suffers from the limitations of the receptive field, and Transformer starts to be acting outstandingly in SR tasks due to its superior remote modeling capability.

IPT~\cite{chen2021pre} is a pre-trained Transformer model on the low-level visual task that introduces the Transformer module in the feature map processing stage.  HAT~\cite{chen2022activating} is a Hybrid Attention Transformer that combines multiple attention mechanisms, such as channel attention, self-attention, and overlapping cross-attention. SwinIR~\cite{liang2021swinir} is an image restoration model based on Swin Transformer. However, the potential of the Transformer still cannot be fully exploited by existing work, and our method adapts the SwinIR-based network architecture by introducing the SFB module based on FFC~\cite{chi2020fast} that can activate more input information. 
LaMa~\cite{suvorov2022resolution} proposes a new image restoration network based on FFC. Inspired by LaMa, we propose the SFB, which employs large receptive field operations within early layers of the network, and can take advantage of long dependencies to use more pixels for better performance.


\begin{figure}[t]
	\centering
	\includegraphics[width=1.0\linewidth]{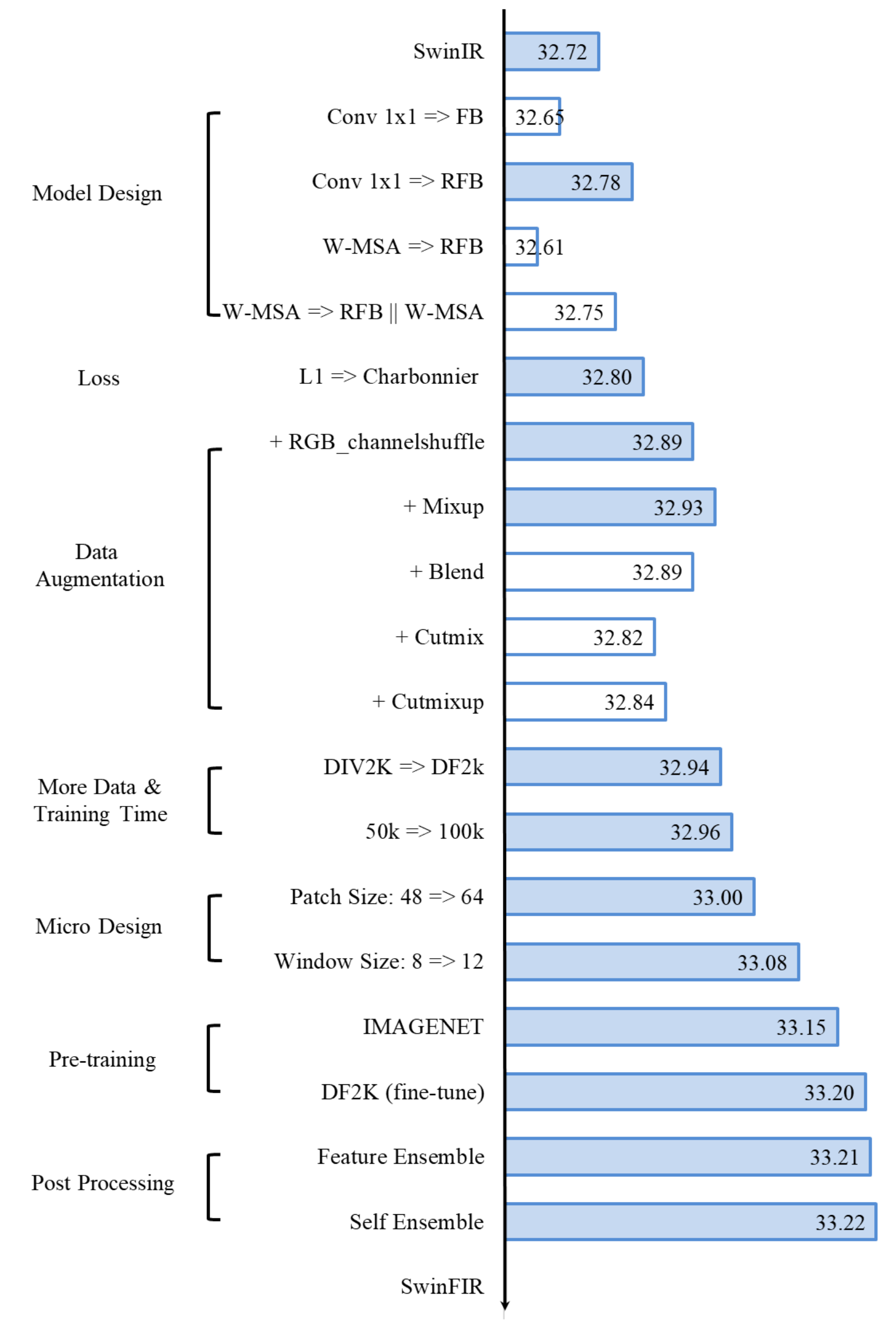}
	\vspace{-5mm}
	\caption{The evolution trajectory from SwinIR to SwinFIR. All models are evaluated on Set5 ($\times$4) dataset.} 
	\label{fig:roadmap}
	\vspace{-3mm}
\end{figure}

\begin{figure*}[t]
	\centering
	\includegraphics[width=1.0\linewidth]{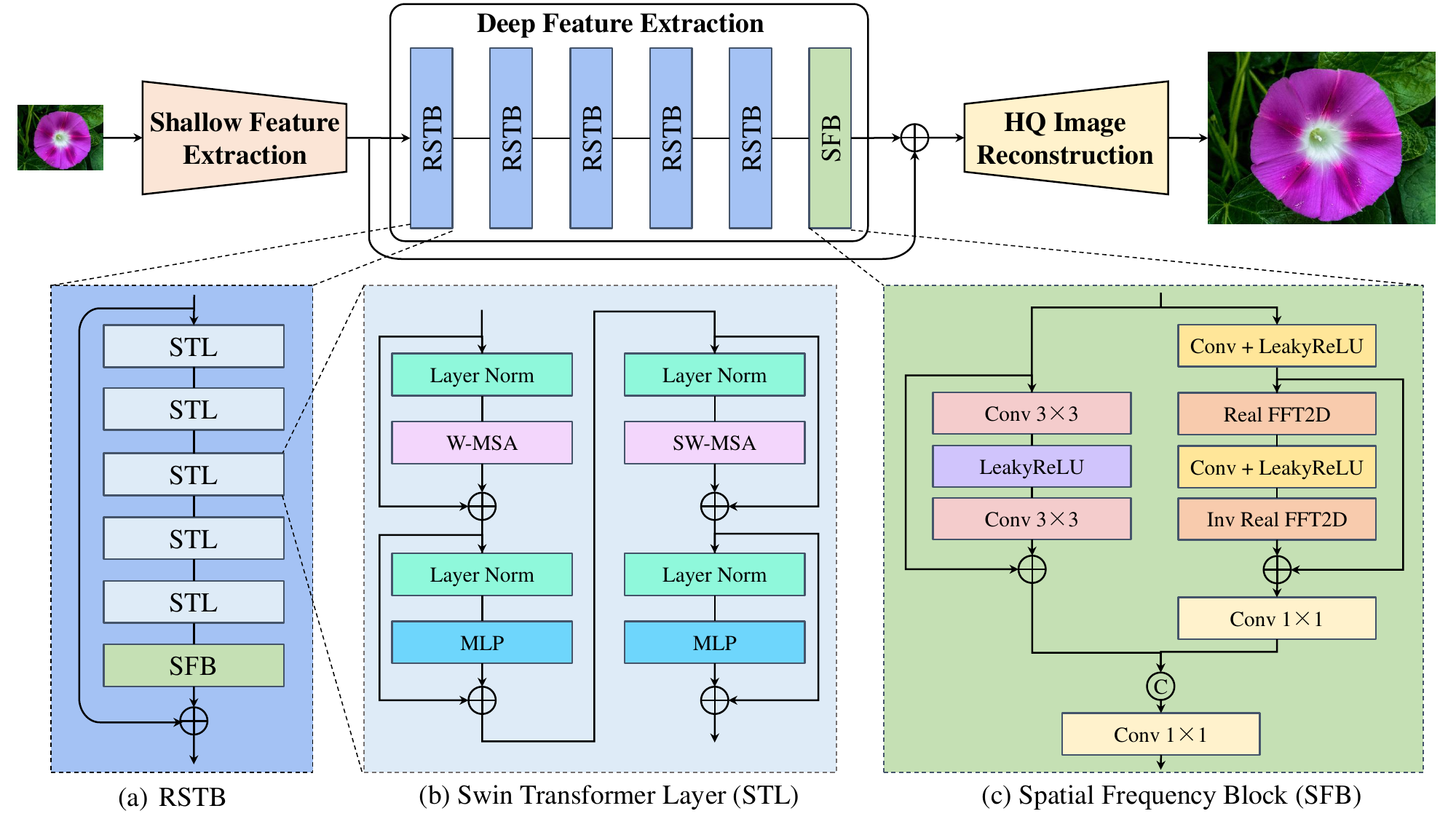}
	\vspace{-3mm}
	\caption{The network architecture of SwinFIR.} 
	\label{fig:SwinFIR_architecture}
	\vspace{-3mm}
\end{figure*}

\section{Methodology}
In this paper, we revisit the strategies for improving image super-resolution, that involve little or no additional model parameters and calculations. The evolution trajectory from SwinIR to SwinFIR is shown in Figure~\ref{fig:roadmap}. LAM~\cite{gu2021interpreting} demonstrate that global information is essential for image super-resolution (SR) since it can activate more pixels and is beneficial to improve the image reconstruction performance. Consequently, we first revisit the SwinIR architecture and introduce the Spatial Frequency Block (SFB) specifically designed for utilizing global information in SR tasks. Then we replace L1 Loss with a more stable Charbonnier Loss~\cite{lai2018fast}. We also revisit a number of data augmentation techniques that can enhance the effectiveness of image super-resolution. 
Loss function and data augmentation strategy are the training and testing free methods. 
We also examine various popular methods for enhancing image super-resolution performance, such as using more training data, enlarging the window size of Swin Transformer and employing pre-training model. Finally, inspired by self-ensemble, we propose a novel post-processing technique, named feature ensemble, to improve the stability of the model without lengthening the training and testing periods. All models are evaluated on Set5 ($\times$4) dataset in Figure~\ref{fig:roadmap}.

\subsection{Model Design}
\begin{figure}[t]
	\centering
	\includegraphics[width=1.0\linewidth]{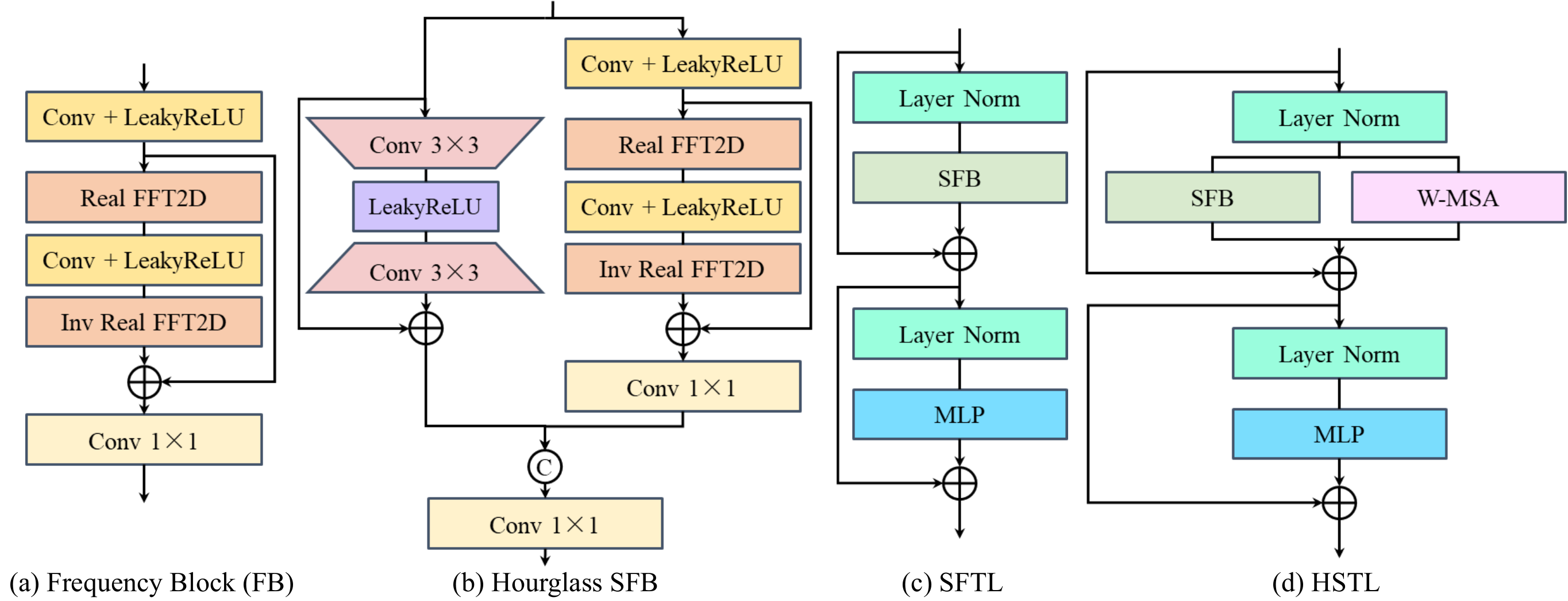}
	\vspace{-5mm}
	\caption{(a) Frequency Block (FB), (b) Hourglass Spatial Frequency Block (Hourglass SFB), (c) Spatial-Frequency Transformer Layer (SFTL) and (d) Hybrid Swin Transformer Layer (HSTL). The detail information in the sec.~\ref{sec:RFB}. } 
	\label{fig:SFB_Block}
	\vspace{-3mm}
\end{figure}

Inspired by SwinIR, we propose SwinFIR using Swin Transformer and Fast Fourier Convolution, as shown in Figure~\ref{fig:SwinFIR_architecture}. SwinFIR consists of three modules: shallow feature extraction, deep feature extraction and high-quality (HQ) image reconstruction modules. The shallow feature extraction and high-quality (HQ) image reconstruction modules adopt the same configuration as SwinIR. The residual Swin Transformer block (RSTB) is a residual block with Swin Transformer layers (STL) and convolutional layers in SwinIR. They all have local receptive fields and cannot extract the global information of the input image. The Fast Fourier Convolution has the ability to extract global features, so we replace the convolution (3$\times$3) with Fast Fourier Convolution and a residual module to fuse global and local features, named Spatial Frequency Block (SFB), to improve the representation ability of model. The analysis of Frequency Block (FB), Spatial-Frequency Transformer Layer (SFTL) and Hybrid Swin Transformer Layer (HSTL) are in Figure~\ref{fig:SFB_Block} and sec.~\ref{sec:RFB}.

The SFB network architecture is shown in Figure~\ref{fig:SwinFIR_architecture}(c) and is composed of two primary components: a spatial conventional convolution operation on the left and a Fast Fourier convolution (FFC) on the right. We concatenate the left and right outputs, and perform a convolution operation to obtain the final result. The formula is as follows,
\begin{align}
	X_{SFB} = H_{SFB}(X)
\end{align}
where the $X$ is the feature map from STL. $H_{SFB}(\cdot)$ represents the SFB module and $X_{SFB}$ is the output feature map after various operations of SFB. We send $X$ into two distinct domains, $X_{spatial}$ and $X_{frequency}$. $X_{spatial}$ is utilized in the spatial domain, and $X_{frequency}$ is intended to capture the long-range context in the frequency domain,
\begin{align}
	& X_{spatial} = H_{spatial}(X) \\
	& X_{frequency}= H_{frequency}(X)
\end{align}
where $H_{spatial}(\cdot)$ is the spatial convolution module and $H_{frequency}(\cdot)$ represents the frequency FFC module. The left spatial convolution module is a residual module for classical SR and a hourglass residual module for lightweight SR, as shown in~\ref{fig:SwinFIR_architecture}(c) and~\ref{fig:SFB_Block}(b) respectively. Compared to a single-layer convolution, we insert a residual connection and convolution layer to increase the expressiveness of the model. Experiments have shown that this simple modification increases performance dramatically. The $X_{spatial}$ is also represented as,
\begin{align}
	X_{spatial} = H_{CLC}(X) + X
\end{align}
where $H_{CLC}(\cdot)$ denotes a 3$\times$3 convolution layer at the head and tail, and LeakyReLU operation is conducted between convolution layers. In the right frequency module, we transform the conventional spatial features into the frequency domain to extract the global information by using the 2-D Fast Fourier Transform (FFT). And we then perform inverse 2-D FFT operation to obtain spatial domain features. The $X_{frequency}$ is also represented as,
\begin{align}
	& X = H_{CL}(X) \\
	& X_{frequency} = H_{C}(H_{FLF}(X) + X)
\end{align}
where $H_{CL}(\cdot)$ denotes a convolution layer and LeakyReLU. $H_{FLF}(\cdot)$ contains the following series of operations, a 2-D FFT based on a channel-wise, a compilation operation with frequency convolution and LeakeyReLU and an inverse 2-D FFT operation. The number of channels is then reduced in half by a convolution operation,
\begin{align}
	X_{SFB}=H_{C}([X_{spatial}|| X_{frequency}])
\end{align}
where $H_{C}(\cdot)$ denotes a convolution layer and $||$ stands for the concatenation operator.

\subsection{Loss Function}
In addition to the structure of the neural network, the loss function also determines whether the model can achieve good results. In low-level visual tasks, such as super resolution and deblurring, the $L2$~\cite{dong2016accelerating}, $L1$~\cite{zhang2018image}, perceptual and adversarial~\cite{sajjadi2017enhancenet} loss functions are often used to optimize neural networks. However, we use Charbonnier loss function~\cite{lai2018fast} to optimize our SwinFIR to get better performance than other loss functions. In the training phase, the loss function is minimized by training data $\{I_L^i, I_H^i\}_{i=1}^N$ to update the parameters, $N$ represent the numbers of training images. The Charbonnier loss function is,
\begin{align}
	L(\theta) = \frac{1}{N} \sum_{i=1}^N \sqrt{(SwinFIR(I_L^i, \theta) - I_H^i)^2 + \varepsilon}
	\label{eq:loss_function}
\end{align}
where $\theta$ denotes the parameters of SwinFIR.

\subsection{Data Augmentation}
Radu~\etal propose rotation and flip data enhancement approaches based on spatial transformation, which is widely used at low-level tasks. However, data augmentation based on the pixel-domain, which is extensively used and has yielded impressive results in high-level tasks, is rarely studied in low-level tasks. In this paper, in addition to flip and rotation, we revisit the effect of data augmentations based on the pixel-domain on image super-resolution, such as RGB channel shuffle, Mixup, Blend, CutMix and CutMixup. RGB channel shuffle randomly shuffles the RGB channels of input images for color augmentation. Mixup randomly mixes the two images according to a certain proportion. Blend randomly adds fixed pixel to input images. CutMix and CutMixup are the combination of Mixup and Cutout. We illustrate in Figure~\ref{fig:roadmap} how various data augmentations affect the performance of image super-resolution on the Set5 dataset. All techniques, except CutMix and CutMixup which destroy visual continuity, are used for data augmentation and achieved performance gains.

Using more training data, enlarging the window size of Swin Transformer and employing pre-training model have all been demonstrated to be feasible in previous studies, so we won't discuss these here.

\begin{figure}[t]
	\centering
	\includegraphics[width=1.0\linewidth]{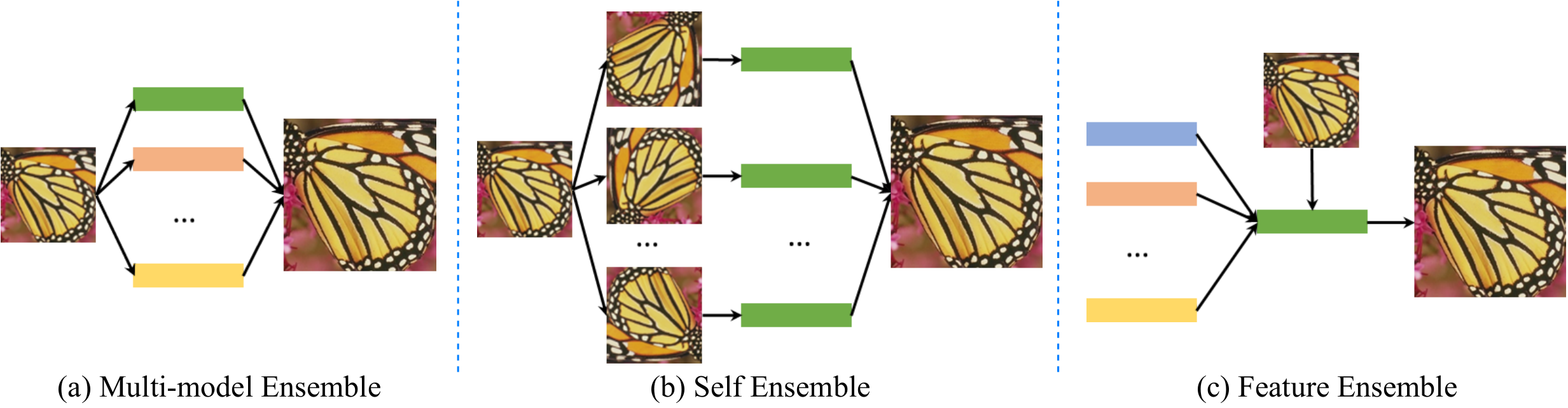}
	\vspace{-6mm}
	\caption{(a) Multi-model Ensemble, (b) Self Ensemble and (c) Feature Ensemble. Rectangles with different colors represent different model parameters.} 
	\label{fig:FeatureEnsemble}
	\vspace{-5mm}
\end{figure}

\subsection{Feature Ensemble}
From the beginning of training the model to the convergence, there will be many intermediate models. In general, the model with the highest performance on the validation set will be selected as the final one, and other models will be deleted. Multi-model ensemble and self-ensembles are often used to improve the performance of SR, as shown in Figure~\ref{fig:FeatureEnsemble}. Multi-model ensemble combines the inference results from various models. Self-ensemble averages the transformed outputs from one model and input image. They have the same drawback, that is, they will multiply the inference time. We propose a novel ensemble strategy without lengthening the training and testing periods. Specifically, we select multiple models that performed well on the validation dataset and combine them using the weighted average method. Our feature ensemble strategy can steadily improve the performance of the model and can be applied to any task, including low-level and high-level.
\begin{align}
	SwinFIR(\theta) = \sum_{i=1}^n {SwinFIR(\theta)^i * \alpha^i}
\end{align}
where $\theta$ denotes the parameter sets of SwinFIR, $n$ is the numbers of models. $\alpha$ is the weight of each model and the $\alpha = \frac{1}{n}$ in this paper. 

\section{Experiments}

\begin{figure*}[t]
	\centering
	\includegraphics[width=1.0\linewidth]{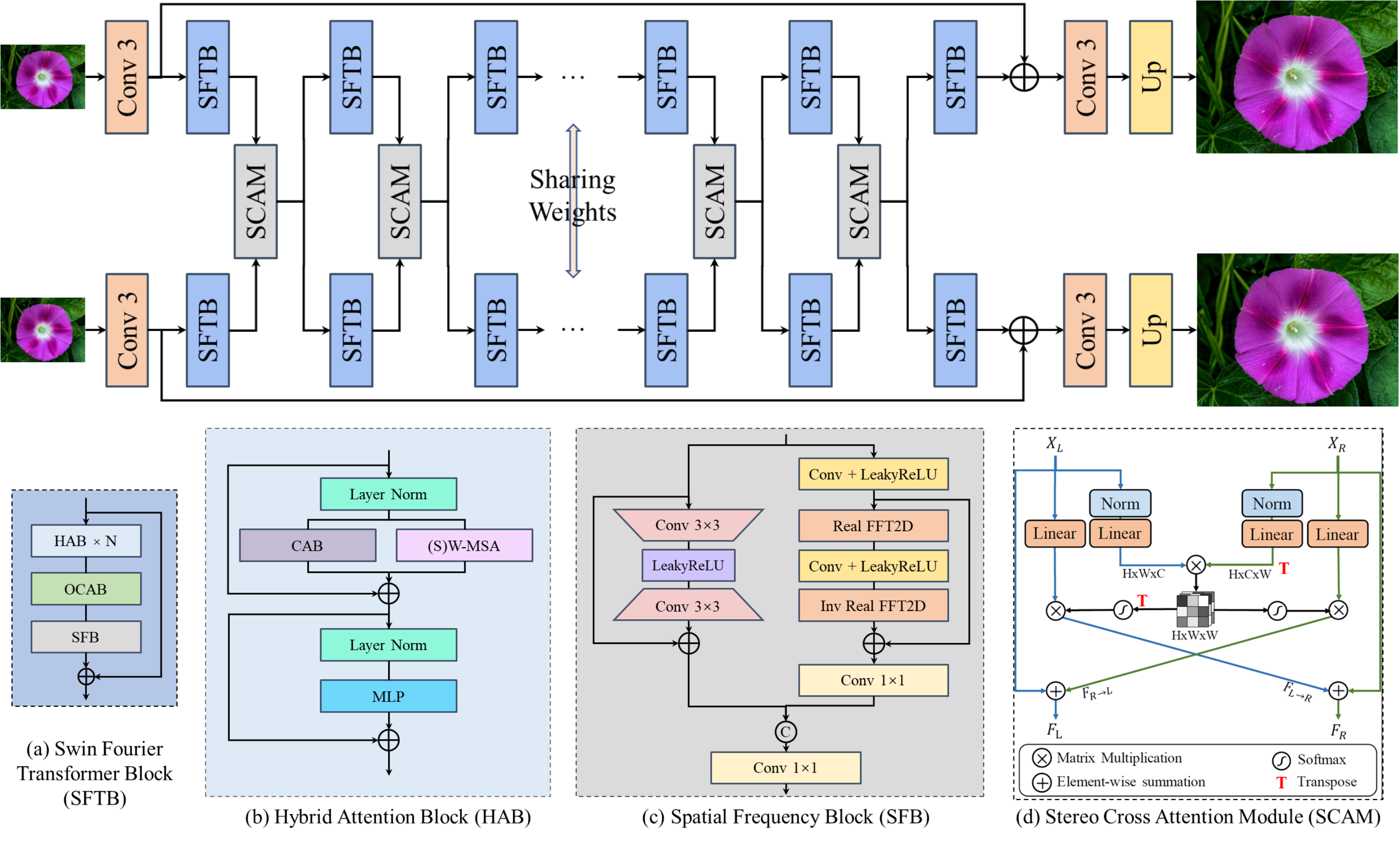}
	\caption{\small
		The network architecture of our SwinFIRSSR for stereo image super-resolution.
	}
	\label{fig:SwinFIRSSR}
\end{figure*}

\begin{figure}[htb]
	\centering
	\includegraphics[width=1.0\linewidth]{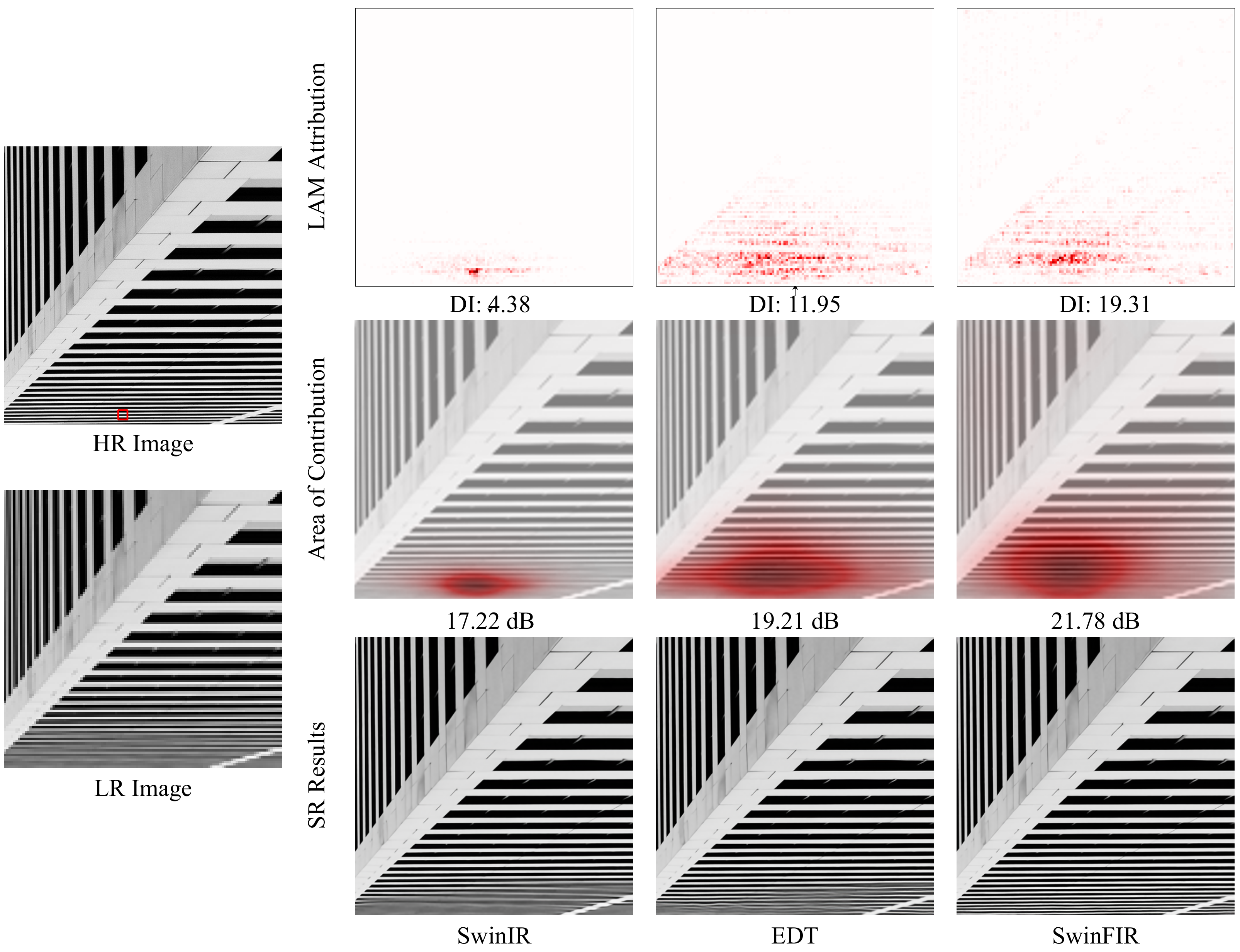}
	\caption{LAM comparisons between SwinIR and EDT on Urban100($\times4$) img\_011. The results indicate that SwinFIR utilize more information than SwinIR and EDT.} 
	\label{fig:LAM}
\end{figure}

\begin{table*}[!t]
	\centering
		\scalebox{0.9}
		{
		\begin{tabular}{lcccccccccccc}
			\hline
			\multirow{2}{*}{Method} & \multirow{2}{*}{Scale} & \multirow{2}{*}{\makecell{Training\\Dataset}} &
			\multicolumn{2}{c}{Set5} &  \multicolumn{2}{c}{Set14} &  \multicolumn{2}{c}{BSD100} &  \multicolumn{2}{c}{Urban100} &  \multicolumn{2}{c}{Manga109}  
			\\ 
			\cline{4-13}
			&  &  & PSNR & SSIM & PSNR & SSIM & PSNR & SSIM & PSNR & SSIM & PSNR & SSIM 
			\\ 
			\hline
			\hline
			EDSR & $\times$2 & DIV2K %
			& 38.11
			& 0.9602
			& 33.92
			& 0.9195
			& 32.32
			& 0.9013
			& 32.93
			& 0.9351
			& 39.10
			& 0.9773
			\\
			RCAN & $\times$2 & DIV2K %
			& 38.27
			& 0.9614
			& 34.12
			& 0.9216
			& 32.41
			& 0.9027
			& 33.34
			& 0.9384
			& 39.44
			& 0.9786
			\\  
			SAN & $\times$2 & DIV2K %
			& {38.31}
			& {0.9620}
			& {34.07}
			& {0.9213}
			& {32.42}
			& {0.9028}
			& {33.10}
			& {0.9370}
			& {39.32}
			& {0.9792}\\
			IGNN & $\times$2 & DIV2K %
			& {38.24}
			& {0.9613}
			& {34.07}
			& {0.9217}
			& {32.41}
			& {0.9025}
			& {33.23}
			& {0.9383}
			& {39.35}
			& {0.9786}
			\\
			HAN & $\times$2 & DIV2K %
			& {38.27}
			& {0.9614}
			& {34.16}
			& {0.9217}
			& {32.41}
			& {0.9027}
			& {33.35}
			& {0.9385}
			& {39.46}
			& {0.9785}              
			\\
			NLSN & $\times$2 & DIV2K %
			& 38.34 
			& 0.9618 
			& 34.08 
			& 0.9231
			& 32.43 
			& 0.9027 
			& 33.42
			& 0.9394
			& 39.59
			& 0.9789
			\\
			SwinIR & $\times$2 & DF2K %
			& 38.42
			& 0.9623
			& 34.46
			& 0.9250
			& 32.53
			& 0.9041
			& 33.81
			& 0.9427
			& 39.92
			& 0.9797
			\\
			EDT & $\times$2 & DF2K %
			& 38.45
			& 0.9624
			& 34.57
			& 0.9258
			& 32.52
			& 0.9041
			& 33.80
			& 0.9425
			& 39.93
			& 0.9800
			\\
			\textbf{SwinFIR} (Ours) & $\times$2 & DF2K %
			& 38.57
			& 0.9630
			& 34.66
			& 0.9263
			& 32.59
			& 0.9049
			& \textcolor{blue}{34.30}
			& \textcolor{blue}{0.9459}
			& 40.30
			& 0.9809
			\\
			\hdashline
			IPT$^\dagger$ & $\times$2 & ImageNet %
			& {38.37}
			& {-}
			& {34.43}
			& {-}
			& {32.48}
			& {-}
			& {33.76}
			& {-}
			& {-}
			& {-}
			\\
			EDT$^\dagger$ & $\times$2 & DF2K %
			& \textcolor{blue}{38.63}
			& \textcolor{blue}{0.9632}
			& \textcolor{blue}{34.80}
			& \textcolor{blue}{0.9273}
			& \textcolor{blue}{32.62}
			& \textcolor{blue}{0.9052}
			& 34.27
			& 0.9456
			& \textcolor{blue}{40.37}
			& \textcolor{blue}{0.9811}
			\\
			\textbf{SwinFIR}$^\dagger$ (Ours) & $\times$2 & DF2K %
			& \textcolor{red}{38.65}
			& \textcolor{red}{0.9633}
			& \textcolor{red}{34.93}
			& \textcolor{red}{0.9276}
			& \textcolor{red}{32.64}
			& \textcolor{red}{0.9054}
			& \textcolor{red}{34.57}
			& \textcolor{red}{0.9473}
			& \textcolor{red}{40.61}
			& \textcolor{red}{0.9816}
			\\
			\hline
			\hline
			EDSR & $\times$3 & DIV2K %
			& 34.65
			& 0.9280
			& 30.52
			& 0.8462
			& 29.25
			& 0.8093
			& 28.80
			& 0.8653
			& 34.17
			& 0.9476
			\\
			RCAN & $\times$3 & DIV2K %
			& 34.74
			& 0.9299
			& 30.65
			& 0.8482
			& 29.32
			& 0.8111
			& 29.09
			& 0.8702
			& 34.44
			& 0.9499
			\\
			SAN & $\times$3 & DIV2K %
			& {34.75}
			& {0.9300}
			& {30.59}
			& {0.8476}
			& {29.33}
			& {0.8112}
			& {28.93}
			& {0.8671}
			& {34.30}
			& {0.9494}
			\\
			IGNN & $\times$3 & DIV2K %
			& {34.72}
			& {0.9298}
			& {30.66}
			& {0.8484}
			& {29.31}
			& {0.8105}
			& {29.03}
			& {0.8696}
			& {34.39}
			& {0.9496}
			\\
			HAN  & $\times$3 & DIV2K %
			& {34.75}
			& {0.9299}
			& {30.67}
			& {0.8483}
			& {29.32}
			& {0.8110}
			& {29.10}
			& {0.8705}
			& {34.48}
			& {0.9500}
			\\
			NLSN & $\times$3 & DIV2K %
			& 34.85 
			& 0.9306 
			& 30.70 
			& 0.8485 
			& 29.34 
			& 0.8117 
			& {29.25}
			& {0.8726}
			& 34.57 
			& 0.9508  
			\\
			SwinIR & $\times$3 & DF2K %
			& 34.97
			& 0.9318
			& 30.93
			& 0.8534
			& 29.46
			& 0.8145
			& 29.75
			& 0.8826
			& 35.12
			& 0.9537
			\\
			EDT & $\times$3 & DF2K %
			& 34.97
			& 0.9316
			& 30.89
			& 0.8527
			& 29.44
			& 0.8142
			& 29.72
			& 0.8814
			& 35.13
			& 0.9534
			\\
			\textbf{SwinFIR} (Ours) & $\times$3 & DF2K %
			& \textcolor{blue}{35.13}
			& \textcolor{blue}{0.9328}
			& \textcolor{blue}{31.13}
			& \textcolor{blue}{0.8556}
			& 29.52
			& 0.8161
			& \textcolor{blue}{30.20}
			& \textcolor{blue}{0.8885}
			& \textcolor{blue}{35.53}
			& \textcolor{blue}{0.9554}
			\\
			\hdashline
			IPT$^\dagger$ & $\times$3 & ImageNet %
			& {34.81}
			& {-}
			& {30.85}
			& {-}
			& {29.38}
			& {-}
			& {29.49}
			& {-}
			& {-}
			& {-}
			\\
			EDT$^\dagger$ & $\times$3 & DF2K %
			& \textcolor{blue}{35.13}
			& \textcolor{blue}{0.9328}
			& 31.09
			& 0.8553
			& \textcolor{blue}{29.53}
			& \textcolor{blue}{0.8165}
			& 30.07
			& 0.8863
			& 35.47
			& 0.9550
			\\
			\textbf{SwinFIR}$^\dagger$ (Ours) & $\times$3 & DF2K %
			& \textcolor{red}{35.15}
			& \textcolor{red}{0.9330}
			& \textcolor{red}{31.24}
			& \textcolor{red}{0.8566}
			& \textcolor{red}{29.55}
			& \textcolor{red}{0.8169}
			& \textcolor{red}{30.43}
			& \textcolor{red}{0.8913}
			& \textcolor{red}{35.77}
			& \textcolor{red}{0.9563}
			\\
			\hline
			\hline
			EDSR & $\times$4 & DIV2K %
			& 32.46
			& 0.8968
			& 28.80
			& 0.7876
			& 27.71
			& 0.7420
			& 26.64
			& 0.8033
			& 31.02
			& 0.9148
			\\
			RCAN & $\times$4 & DIV2K %
			& 32.63
			& 0.9002
			& 28.87
			& 0.7889
			& 27.77
			& 0.7436
			& 26.82
			& 0.8087
			& 31.22
			& 0.9173
			\\
			SAN & $\times$4 & DIV2K %
			& {32.64}
			& {0.9003}
			& {28.92}
			& {0.7888}
			& {27.78}
			& {0.7436}
			& {26.79}
			& {0.8068}
			& {31.18}
			& {0.9169}
			\\
			IGNN & $\times$4 & DIV2K %
			& {32.57}
			& {0.8998}
			& {28.85}
			& {0.7891}
			& {27.77}
			& {0.7434}
			& {26.84}
			& {0.8090}
			& {31.28}
			& {0.9182}
			\\
			HAN & $\times$4 & DIV2K %
			& {32.64}
			& {0.9002}
			& {28.90}
			& {0.7890}
			& {27.80}
			& {0.7442}
			& {26.85}
			& {0.8094}
			& {31.42}
			& {0.9177}
			\\
			NLSN & $\times$4 & DIV2K %
			& 32.59 
			& 0.9000 
			& 28.87 
			& 0.7891 
			& 27.78 
			& 0.7444 
			& {26.96}
			& {0.8109}
			& 31.27 
			& 0.9184
			\\
			SwinIR & $\times$4 & DF2K %
			& 32.92
			& 0.9044
			& 29.09
			& 0.7950
			& 27.92
			& 0.7489
			& 27.45
			& 0.8254
			& 32.03
			& 0.9260
			\\
			EDT & $\times$4 & DF2K %
			& 32.82
			& 0.9031
			& 29.09
			& 0.7939
			& 27.91
			& 0.7483
			& 27.46
			& 0.8246
			& 32.05
			& 0.9254
			\\
			\textbf{SwinFIR} (Ours) & $\times$4 & DF2K %
			& \textcolor{blue}{33.08}
			& 0.9048
			& 29.21
			& \textcolor{blue}{0.7971}
			& 27.98
			& 0.7508
			& \textcolor{blue}{27.87}
			& \textcolor{blue}{0.9348}
			& \textcolor{blue}{32.52}
			& \textcolor{blue}{0.9292}
			\\
			\hdashline
			IPT$^\dagger$ & $\times$4 & ImageNet %
			& {32.64}
			& {-}
			& {29.01}
			& {-}
			& {27.82}
			& {-}
			& {27.26}
			& {-}
			& {-}
			& {-}
			\\
			EDT$^\dagger$ & $\times$4 & DF2K %
			& 33.06
			& \textcolor{blue}{0.9055}
			& \textcolor{blue}{29.23}
			& \textcolor{blue}{0.7971}
			& \textcolor{blue}{27.99}
			& \textcolor{blue}{0.7510}
			& 27.75
			& 0.8317
			& 32.39
			& 0.9283
			\\
			\textbf{SwinFIR}$^\dagger$ (Ours) & $\times$4 & DF2K %
			& \textcolor{red}{33.20}
			& \textcolor{red}{0.9068}
			& \textcolor{red}{29.36}
			& \textcolor{red}{0.7993}
			& \textcolor{red}{28.03}
			& \textcolor{red}{0.7520}
			& \textcolor{red}{28.12}
			& \textcolor{red}{0.8393}
			& \textcolor{red}{32.83}
			& \textcolor{red}{0.9314}
			\\
			\hline             
		\end{tabular}}
		\caption{Quantitative comparison with state-of-the-art methods on benchmark datasets on the Y channel from the YCbCr space for \textbf{classical image SR}. The top two results are marked in \textcolor{red}{red} and \textcolor{blue}{blue}. ``$\dagger$'' indicates that methods adopt pre-training strategy on ImageNet.}
		\label{tab:SwinFIR_quantitative_results}
\end{table*}

\begin{figure*}[t]
	\scriptsize
	\centering
	\begin{tabular}{cc}
		\hspace{-0.4cm}
		\begin{adjustbox}{valign=t}
			\begin{tabular}{c}
				\includegraphics[width=0.236\textwidth]{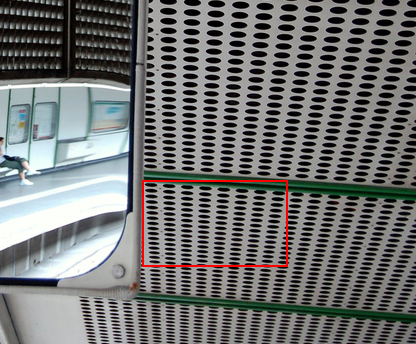}
				\\
				Urban100($\times$4): img\_004
			\end{tabular}
		\end{adjustbox}
		\hspace{-0.46cm}
		\begin{adjustbox}{valign=t}
			\begin{tabular}{cccccc}
				\includegraphics[width=0.147\textwidth]{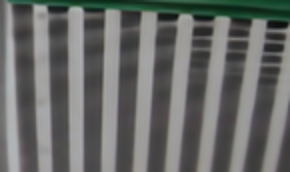} \hspace{-4mm} &
				\includegraphics[width=0.147\textwidth]{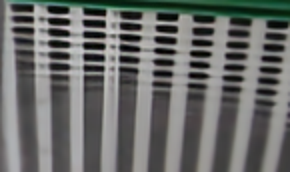} \hspace{-4mm} &
				\includegraphics[width=0.147\textwidth]{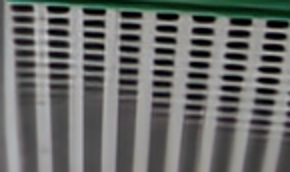} \hspace{-4mm} &
				\includegraphics[width=0.147\textwidth]{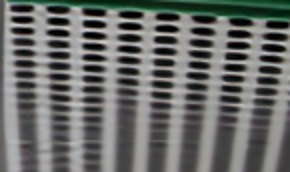} \hspace{-4mm} &
				\includegraphics[width=0.147\textwidth]{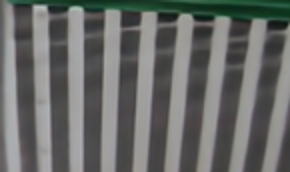} \hspace{-4mm} 
				
				\\
				EDSR \hspace{-4mm} &
				IGNN \hspace{-4mm} &
				HAN \hspace{-4mm} &
				IPT \hspace{-4mm} &
				RNAN \hspace{-4mm}
				\\

				\includegraphics[width=0.147\textwidth]{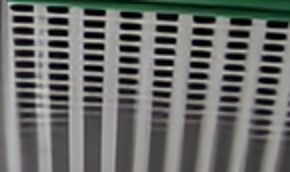} \hspace{-4mm} &
				\includegraphics[width=0.147\textwidth]{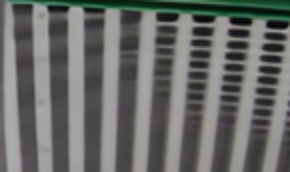} \hspace{-4mm} &
				\includegraphics[width=0.147\textwidth]{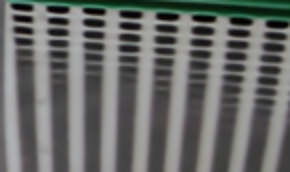} \hspace{-4mm}   &
				\includegraphics[width=0.147\textwidth]{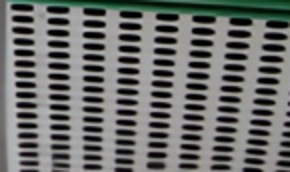} \hspace{-4mm} &
				\includegraphics[width=0.147\textwidth]{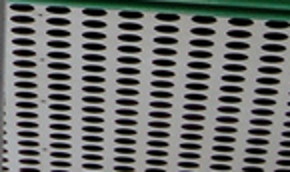} \hspace{-4mm} 
				\\ 
				RCAN \hspace{-4mm} &
				SwinIR  \hspace{-4mm} &
				EDT \hspace{-4mm} &
				SwinFIR(ours) \hspace{-4mm} &
				Reference \hspace{-4mm}
				\\
			\end{tabular}
		\end{adjustbox}
		\vspace{1mm}
		\\
		
		 \hspace{-0.4cm}
		 \begin{adjustbox}{valign=t}
		 	\begin{tabular}{c}
		 		\includegraphics[width=0.236\textwidth]{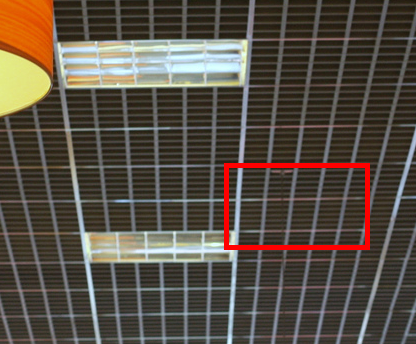}
		 		\\
		 		Urban100($\times$4): img\_044
		 	\end{tabular}
		 \end{adjustbox}
		 \hspace{-0.46cm}
		 \begin{adjustbox}{valign=t}
		 	\begin{tabular}{cccccc}
		 		\includegraphics[width=0.147\textwidth]{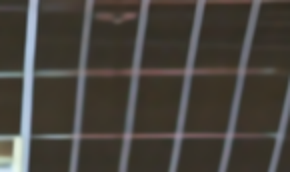} \hspace{-4mm} &
		 		\includegraphics[width=0.147\textwidth]{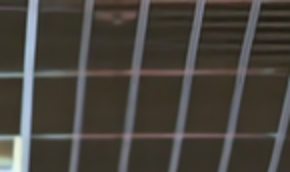} \hspace{-4mm} &
		 		\includegraphics[width=0.147\textwidth]{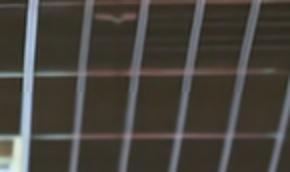} \hspace{-4mm} &
		 		\includegraphics[width=0.147\textwidth]{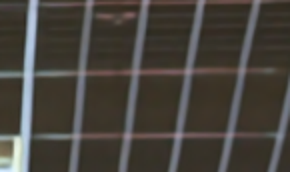} \hspace{-4mm} &
		 		\includegraphics[width=0.147\textwidth]{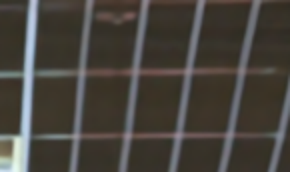} \hspace{-4mm} 
				
		 		\\
		 		EDSR \hspace{-4mm} &
		 		IGNN \hspace{-4mm} &
		 		HAN \hspace{-4mm} &
		 		IPT \hspace{-4mm} &
		 		RNAN \hspace{-4mm}
		 		\\

		 		\includegraphics[width=0.147\textwidth]{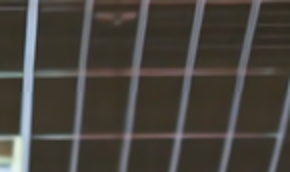} \hspace{-4mm} &
		 		\includegraphics[width=0.147\textwidth]{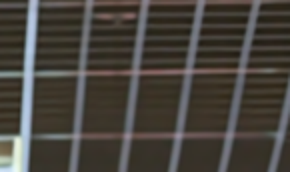} \hspace{-4mm} &
		 		\includegraphics[width=0.147\textwidth]{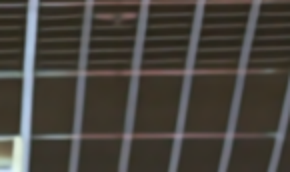} \hspace{-4mm}   &
		 		\includegraphics[width=0.147\textwidth]{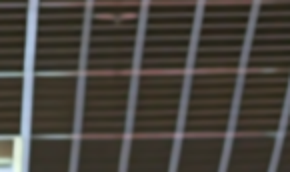} \hspace{-4mm} &
		 		\includegraphics[width=0.147\textwidth]{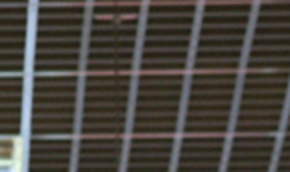} \hspace{-4mm} 
		 		\\ 
		 		RCAN \hspace{-4mm} &
		 		SwinIR  \hspace{-4mm} &
		 		EDT \hspace{-4mm} &
		 		SwinFIR(ours) \hspace{-4mm} &
		 		Reference \hspace{-4mm}
		 		\\
		 	\end{tabular}
		 \end{adjustbox}
		 \vspace{1mm}
		 \\
		
		\hspace{-0.4cm}
		\begin{adjustbox}{valign=t}
			\begin{tabular}{c}
				\includegraphics[width=0.236\textwidth]{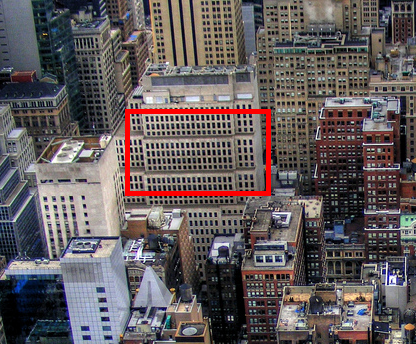}
				\\
				Urban100($\times$4): img\_073
			\end{tabular}
		\end{adjustbox}
		\hspace{-0.46cm}
		\begin{adjustbox}{valign=t}
			\begin{tabular}{cccccc}
				\includegraphics[width=0.147\textwidth]{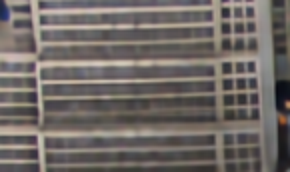} \hspace{-4mm} &
				\includegraphics[width=0.147\textwidth]{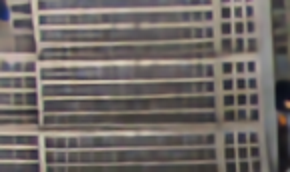} \hspace{-4mm} &
				\includegraphics[width=0.147\textwidth]{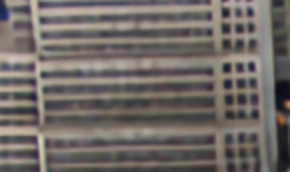} \hspace{-4mm} &
				\includegraphics[width=0.147\textwidth]{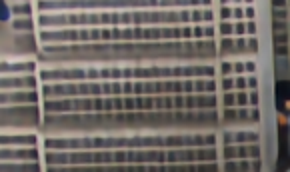} \hspace{-4mm} &
				\includegraphics[width=0.147\textwidth]{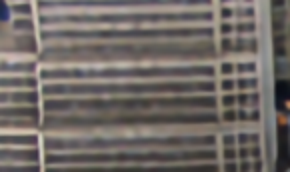} \hspace{-4mm} 
				
				\\
				EDSR \hspace{-4mm} &
				IGNN \hspace{-4mm} &
				HAN \hspace{-4mm} &
				IPT \hspace{-4mm} &
				RNAN \hspace{-4mm}
				\\

				\includegraphics[width=0.147\textwidth]{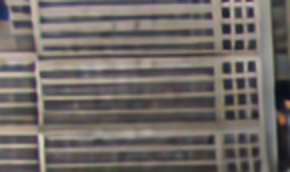} \hspace{-4mm} &
				\includegraphics[width=0.147\textwidth]{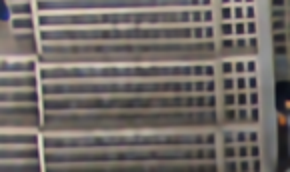} \hspace{-4mm} &
				\includegraphics[width=0.147\textwidth]{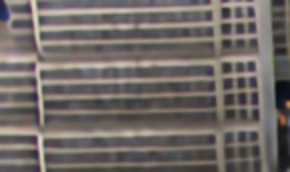} \hspace{-4mm}   &
				\includegraphics[width=0.147\textwidth]{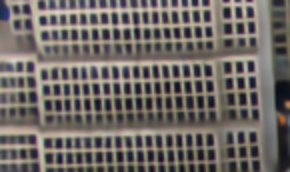} \hspace{-4mm} &
				\includegraphics[width=0.147\textwidth]{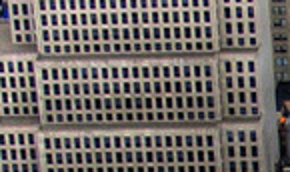} \hspace{-4mm} 
				\\ 
				RCAN \hspace{-4mm} &
				SwinIR  \hspace{-4mm} &
				EDT \hspace{-4mm} &
				SwinFIR(ours) \hspace{-4mm} &
				Reference \hspace{-4mm}
				\\
			\end{tabular}
		\end{adjustbox}
		\vspace{1mm}
		\\
	\end{tabular}
	\caption{Visual results ($\times$4) achieved by different methods on the Urban100 dataset (\textbf{classical image SR}).}
	\label{fig:Urban100_SwinFIR}
\end{figure*}

\subsection{Datasets}
Following EDT~\cite{li2021efficient}, we pre-train the SwinFIR on ImageNet 2012~\cite{deng2009imagenet}. 
The SwinFIR is then fine-tuned for Classical and Lightweight SR using sub-images (384$\times$384) that were generated by cropping the high-resolution DF2K (DIV2K~\cite{lim2017enhanced} + Flicker2K~\cite{timofte2017ntire}) dataset. We perform validation on Image Super-Resolution benchmark datasets Set5, Set14, BSD100, Urban100 and Manga109 for Classical and Lightweight SR. 
We perform data augmentation on training data to improve the robustness of SwinFIR, which includes horizontal flip, vertical flip, rotation, channel shuffle and Mixup.

\subsection{Implementation Details}
We revisit the long-dependent modeling capabilities of SwinIR and propose an efficient global feature extractor based on Fast Fourier Convolution (FFC). Specifically, we replace the convolution layer in RSTB of SwinIR with Spatial Frequency Block (SFB). For classical image SR, we utilize the same configuration as SwinIR.
We also investigate how performance of SR is affected by large window and patch size. As a result, we use a larger window size 12 and patch size 60 in our work. For lightweight image SR, we also decrease RSTB number and channel number to 4 and 60 follow SwinIR, respectively. However, we use 5 STL in the second and third RSTB to accelerate training and inference time.

We extend the SwinFIR to SwinFIRSSR following NAFSSR~\cite{chu2022nafssr} and HAT~\cite{chen2022activating}, and verify the effectiveness of our method in the stereo image super-resolution task, as shown in Figure.~\ref{fig:SwinFIRSSR}. HAT proposed the Residual Hybrid Attention Group (RHAG) to activating more pixel in image super-resolution transformer for improve the performance. RHAG contains N hybrid attention blocks (HAB), an overlapping cross-attention block (OCAB) and a 3$\times$3 convolutional layer, we replace the convolution (3$\times$3) with Fast Fourier Convolution and a residual module to fuse global and local features, named Spatial-frequency Block (SFB), to improve the representation ability of model. We also follow the NAFSSR to attend and fuse the left/right viewpoint features by using stereo cross-attention module (SCAM). 

We use the Adam with $\beta_1=0.9$ and $\beta_2=0.99$ and weight decay 0 by default to optimize the Charbonnier loss function for classical image SR and lightweight image SR. In the pre-training stage, the initial learning rate is 2e-4 and reduces by 50\% in 500,000, 800,000, 900,000, and 950,000 iterations for a total of 1,000,000 iterations, respectively. In the fine-tune stage, the learning rate is decreased to 1e-5 for classical and lightweight image SR. 

\subsection{Comparison to state-of-the-arts methods}
\subsubsection{Classical Image Super-Resolution}
Modern algorithms such as EDSR~\cite{lim2017enhanced}, RCAN~\cite{zhang2018image}, SAN~\cite{dai2019second}, IGNN~\cite{zhou2020cross}, RNAN~\cite{zhang2019residual}, HAN~\cite{niu2020single}, NLSA~\cite{mei2021image}, IPT~\cite{chen2021pre}, SwinIR~\cite{liang2021swinir}, and EDT~\cite{li2021efficient} are compared to our SwinFIR. 
The DIV2K dataset was used to train the CNN-based methods EDSR, RCAN, SAN, IGNN, RNAN, HAN, and NLSA. Networks based on Vision Transformer include IPT, SwinIR, and EDT. IPT and EDT are trained on the ImageNet dataset, while EDT is fine-tuned on the DF2K dataset to get better performance. And SwinIR only is trained on the DF2K dataset. Following EDT, our SwinFIR is first trained on the ImageNet, and then fine-tuned on the DF2K dataset. 
Table~\ref{tab:SwinFIR_quantitative_results} displays the quantitative results on benchmark datasets for classical SR. Our SwinFIR achieves the best SR performance on $\times$2, $\times$3 and $\times$4 scales compared with other state-of-the-art methods. Especially, SwinFIR improves the PSNR of SwinIR from 27.45 dB and 32.03 dB to 28.12 dB and 32.83 dB on $\times$4 scales of Urban100 and Manga109 datasets respectively, 0.77 dB and 0.80 dB higher than its. It demonstrates the effectiveness of our proposed method and represents a major improvement over the image super-resolution task. Even without pre-training, our SwinFIR achieves better or comparable performance than EDT with pre-training, 0.13 dB higher on $\times$4 scales of Manga109 datasets.

Visual results are shown in Figure~\ref{fig:Urban100_SwinFIR}, and the images restored by our SwinFIR are clearer. Our SwinFIR can restore high-frequency details based on Fast Fourier Convolution (FFC). Especially, Our SwinFIR performs better when trying to restore images with periodic transformations. It is worth mentioning that the current approaches, whether based on CNN or Transformer, are inadequate for challenging samples, as shown in Figure~\ref{fig:Urban100_SwinFIR}. And our method addresses this issue by making slight adjustments to the SwinIR, which can significantly improve performance.

The LAM results are shown in Figure~\ref{fig:LAM}. The LAM attribution map and area of contribution accurately depict the significance of each pixel and receptive field size in the input LR image when reconstructing the red box region of the SR images. The Diffusion Index (DI) illustrates the range of relevant and utilised pixels. The higher the DI, the wider range of pixels are used. Furthermore, SwinFIR outperforms SwinIR and EDT in terms of DI, as shown in Figure~\ref{fig:LAM}. And almost all of the pixels from the input LR image are used to restore the SR image in SwinFIR, according to the LAM attribution map. The qualitative and quantitative results demonstrate that SwinIR does has a limited receptive field, and our SwinFIR based FFC takes advantage of long dependencies to use more pixels for better performance.

\subsubsection{Lightweight Image Super-Resolution}
\begin{figure*}[t]
	\scriptsize
	\centering
	\begin{tabular}{cc}
		\hspace{-0.4cm}
		 \begin{adjustbox}{valign=t}
		 	\begin{tabular}{c}
		 		\includegraphics[width=0.236\textwidth]{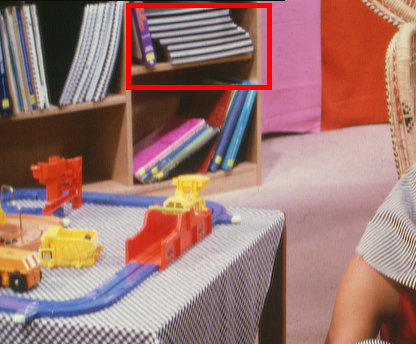}
		 		\\
		 		Set14($\times$4): barbara
		 	\end{tabular}
		 \end{adjustbox}
		 \hspace{-0.46cm}
		 \begin{adjustbox}{valign=t}
		 	\begin{tabular}{cccccc}
		 		\includegraphics[width=0.147\textwidth]{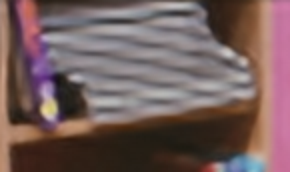} \hspace{-4mm} &
		 		\includegraphics[width=0.147\textwidth]{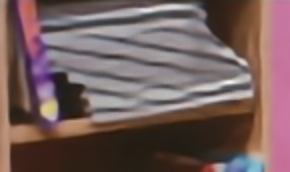} \hspace{-4mm} &
		 		\includegraphics[width=0.147\textwidth]{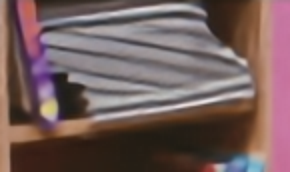} \hspace{-4mm} &
		 		\includegraphics[width=0.147\textwidth]{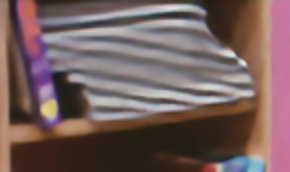} \hspace{-4mm} &
		 		\includegraphics[width=0.147\textwidth]{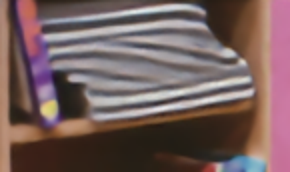} \hspace{-4mm} 
				
		 		\\
		 		SRCNN \hspace{-4mm} &
		 		LapSRN \hspace{-4mm} &
		 		DRRN \hspace{-4mm} &
		 		CARN-M \hspace{-4mm} &
		 		SRFBN-S \hspace{-4mm}
		 		\\

		 		\includegraphics[width=0.147\textwidth]{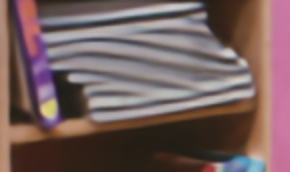} \hspace{-4mm} &
		 		\includegraphics[width=0.147\textwidth]{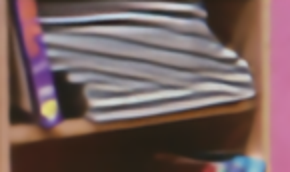} \hspace{-4mm} &
		 		\includegraphics[width=0.147\textwidth]{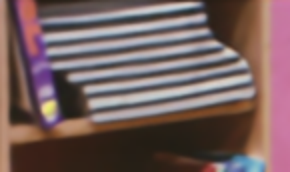} \hspace{-4mm}   &
		 		\includegraphics[width=0.147\textwidth]{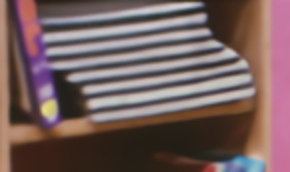} \hspace{-4mm} &
		 		\includegraphics[width=0.147\textwidth]{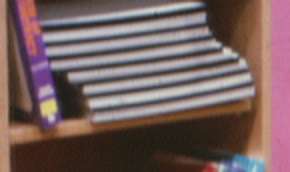} \hspace{-4mm} 
		 		\\ 
		 		IMDN \hspace{-4mm} &
		 		SwinIR  \hspace{-4mm} &
		 		EDT \hspace{-4mm} &
		 		SwinFIR(ours) \hspace{-4mm} &
		 		Reference \hspace{-4mm}
		 		\\
		 	\end{tabular}
		 \end{adjustbox}
		 \vspace{1mm}
		 \\
		
		\hspace{-0.4cm}
		\begin{adjustbox}{valign=t}
			\begin{tabular}{c}
				\includegraphics[width=0.236\textwidth]{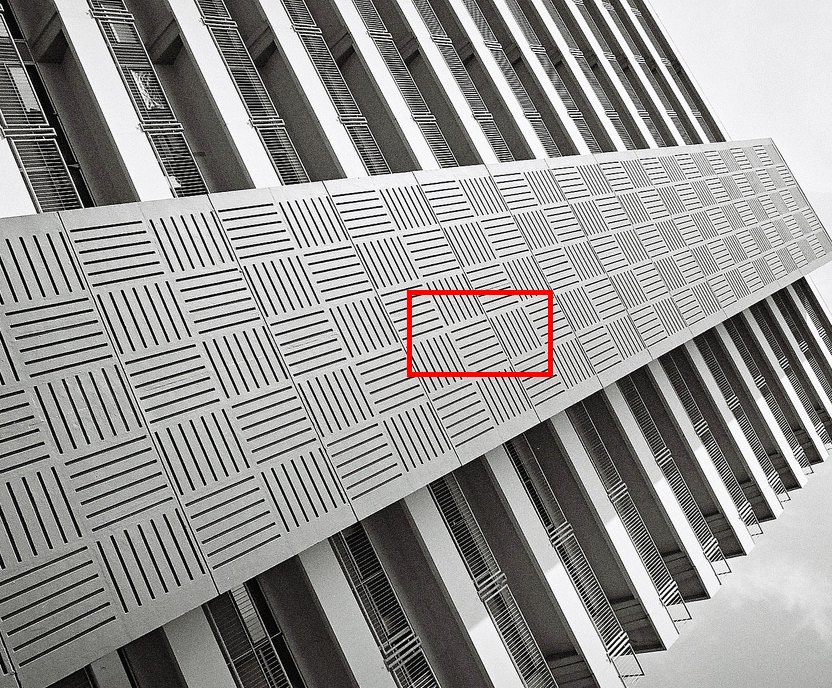}
				\\
				Urban100($\times$4): img\_092
			\end{tabular}
		\end{adjustbox}
		\hspace{-0.46cm}
		\begin{adjustbox}{valign=t}
			\begin{tabular}{cccccc}
				\includegraphics[width=0.147\textwidth]{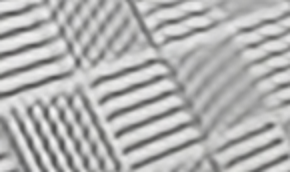} \hspace{-4mm} &
				\includegraphics[width=0.147\textwidth]{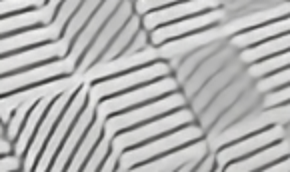} \hspace{-4mm} &
				\includegraphics[width=0.147\textwidth]{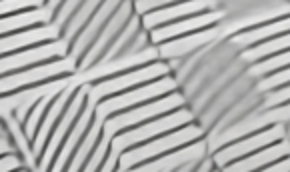} \hspace{-4mm} &
				\includegraphics[width=0.147\textwidth]{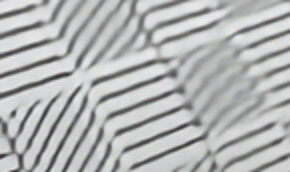} \hspace{-4mm} &
				\includegraphics[width=0.147\textwidth]{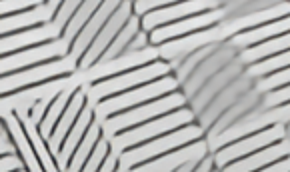} \hspace{-4mm} 
				
				\\
				SRCNN \hspace{-4mm} &
				LapSRN \hspace{-4mm} &
				DRRN \hspace{-4mm} &
				CARN-M \hspace{-4mm} &
				SRFBN-S \hspace{-4mm}
				\\

				\includegraphics[width=0.147\textwidth]{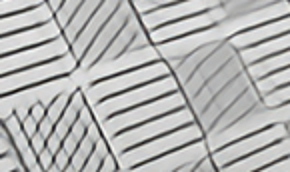} \hspace{-4mm} &
				\includegraphics[width=0.147\textwidth]{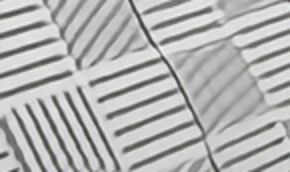} \hspace{-4mm} &
				\includegraphics[width=0.147\textwidth]{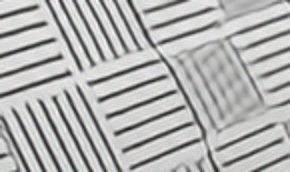} \hspace{-4mm}   &
				\includegraphics[width=0.147\textwidth]{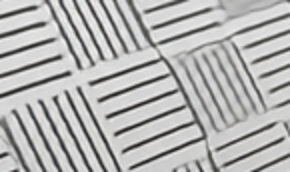} \hspace{-4mm} &
				\includegraphics[width=0.147\textwidth]{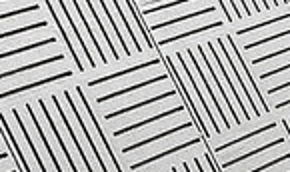} \hspace{-4mm} 
				\\ 
				IMDN \hspace{-4mm} &
				SwinIR  \hspace{-4mm} &
				EDT \hspace{-4mm} &
				SwinFIR(ours) \hspace{-4mm} &
				Reference \hspace{-4mm}
				\\
			\end{tabular}
		\end{adjustbox}
		\vspace{1mm}
		\\
	\end{tabular}
	\vspace{-3mm}
	\caption{Visual results ($\times$4) achieved by different methods on the  Urban100 dataset (\textbf{lightweight image SR}).}
	\label{fig:Urban100_SwinFIR_L}
	\vspace{-3mm}
\end{figure*}

We substitute the Hourglass SFB for SFB in lightweight SwinFIR, named SwinFIR-T. And we compare SwinFIR-T with the SOTA lightweight SR methods, including SRCNN ~\cite{dong2015image}, LapSRN~\cite{lai2017deep}, DRRN~\cite{tai2017image}, CARN-M~\cite{2018Fast}, SRFBN-S~\cite{2019Feedback}, IMDN~\cite{2019Lightweight}, SwinIR (small size) and EDT-T. 
SwinFIR significantly improves image SR performance and achieves the best results for all metrics, as indicated by the quantitative comparison in Table~\ref{tab:SwinFIR_L_quantitative_results}. SwinFIR achieves 31.50dB PSNR on the Manga109 dataset ($\times$4) when the number of parameters are comparable to the SwinIR and EDT-T, which is 0.58 dB and 0.15 dB higher than them, respectively. 
In particular, SwinFIR achieves better or comparable performance to EDT when using a smaller window size. 
Visual results are presented in Figure~\ref{fig:Urban100_SwinFIR_L}, and the images restored by our SwinFIR are sharper and contain more high-frequency detail information. 
EDT and our SwinFIR all can recover the detailed information of the book on Set14 barbara dataset, but the reconstruction results of EDT are more blurring.

\begin{table*}[t]
	\centering
		\scalebox{0.85}{
		\begin{tabular}{ c  l  c  c  c  c  c  c  c  c  c  c  c  c }
			\hline
			\multirow{2}{*}{Scale} & \multirow{2}{*}{Method} & \#Param. & Flops & \multicolumn{2}{c}{Set5} & \multicolumn{2}{c}{Set14} & \multicolumn{2}{c}{BSDS100} & \multicolumn{2}{c}{Urban100} & \multicolumn{2}{c}{Manga109} \\
			\cline{5-14}
			~ & ~ & (K) & (G) & PSNR & SSIM & PSNR & SSIM & PSNR & SSIM & PSNR & SSIM & PSNR & SSIM \\
			\hline
			\hline

			\multirow{5}{*}{$\times 2$} & LAPAR & 548 & 171.5 & 38.01 & 0.9605 & 33.62 & 0.9183 & 32.19 & 0.8999 & 32.10 & 0.9283 & 38.67 & 0.9772 \\
			~ & LatticeNet & 756 & 171.2 & 38.15 & 0.9610 & 33.78 & 0.9193 & 32.25 & 0.9005 & 32.43 & 0.9302 & - & - \\
			~ & SwinIR & 878 & 205.5 & 38.14 & 0.9611 & 33.86 & 0.9206 & 32.31 & 0.9012 & 32.76 & 0.9340 & 39.12 & 0.9783 \\
			~ & EDT-T & 917 & 224.2 & \textcolor{blue}{38.23} & \textcolor{blue}{0.9615} & \textcolor{blue}{33.99} & \textcolor{blue}{0.9209} & \textcolor{blue}{32.37} & \textcolor{blue}{0.9021} & \textcolor{blue}{32.98} & \textcolor{blue}{0.9362} & \textcolor{blue}{39.45} & \textcolor{blue}{0.9789} \\
			~ & \textbf{SwinFIR-T}(Ours) & 872 & 206.3 & \textcolor{red}{38.26} & \textcolor{red}{0.9616} & \textcolor{red}{34.08} & \textcolor{red}{0.9221} & \textcolor{red}{32.38} & \textcolor{red}{0.9024} & \textcolor{red}{33.14} & \textcolor{red}{0.9374} & \textcolor{red}{39.55} & \textcolor{red}{0.9790} \\
			\hline
			
			\multirow{5}{*}{$\times 3$} & LAPAR & 594 & 114.44 & 34.36 & 0.9267 & 30.34 & 0.8421 & 29.11 & 0.8054 & 28.15 & 0.8523 & 33.51 & 0.9441 \\
			~ & LatticeNet & 765 & 77.0 & 34.53 & 0.9281 & 30.39 & 0.8424 & 29.15 & 0.8059 & 28.33 & 0.8538 & - & - \\
			~ & SwinIR & 886 & 93.2 & 34.62 & 0.9289 & 30.54 & 0.8463 & 29.20 & 0.8082 & 28.66 & 0.8624 & 33.98 & 0.9478 \\
			~ & EDT-T & 919 & 103.0 & \textcolor{blue}{34.73} & \textcolor{blue}{0.9299} & \textcolor{blue}{30.66} & \textcolor{blue}{0.8481} & \textcolor{blue}{29.29} & \textcolor{blue}{0.8103} & \textcolor{blue}{28.89} & \textcolor{blue}{0.8674} & \textcolor{blue}{34.44} & \textcolor{blue}{0.9498} \\
			~ & \textbf{SwinFIR-T}(Ours) & 880 & 94.0 & \textcolor{red}{34.75} & \textcolor{red}{0.9300} & \textcolor{red}{30.68} & \textcolor{red}{0.8489} & \textcolor{red}{29.30} & \textcolor{red}{0.8106} & \textcolor{red}{29.04} & \textcolor{red}{0.8697} & \textcolor{red}{34.60} & \textcolor{red}{0.9506} \\					
			\hline
			
			\multirow{5}{*}{$\times 4$} & LAPAR & 659 & 94..8 & 32.15 & 0.8944 & 28.61 & 0.7818 & 27.61 & 0.7366 & 26.14 & 0.7871 & 30.42 & 0.9074 \\
			~ & LatticeNet & 777 & 44.2 & 32.30 & 0.8962 & 28.68 & 0.7830 & 27.62 & 0.7367 & 26.25 & 0.7873 & - & - \\
			~ & SwinIR & 897 & 53.2 & 32.44 &0.8976 & 28.77 & 0.7858 & 27.69 & 0.7406 & 26.47 & 0.7980 & 30.92 & 0.9151 \\
			~ & EDT-T & 922 & 58.5 & \textcolor{blue}{32.53} & \textcolor{blue}{0.8991} & \textcolor{blue}{28.88} & \textcolor{blue}{0.7882} & \textcolor{blue}{27.76} & \textcolor{blue}{0.7433} & \textcolor{blue}{26.71} & \textcolor{blue}{0.8051} & \textcolor{blue}{31.35} & \textcolor{blue}{0.9180} \\
			~ & \textbf{SwinFIR-T}(Ours) & 891 & 54.4 & \textcolor{red}{32.62} & \textcolor{red}{0.9002} & \textcolor{red}{28.95} & \textcolor{red}{0.7898} & \textcolor{red}{27.79} & \textcolor{red}{0.7440} & \textcolor{red}{26.85} & \textcolor{red}{0.8088} & \textcolor{red}{31.50} & \textcolor{red}{0.9199} \\					
			\hline
		\end{tabular}}
		\caption{Quantitative comparison with state-of-the-art methods on benchmark datasets on the Y channel from the YCbCr space for \textbf{lightweight image SR}. The top two results are marked in \textcolor{red}{red} and \textcolor{blue}{blue}.}
		\label{tab:SwinFIR_L_quantitative_results}
\end{table*}

\begin{figure*}[t]
	\scriptsize
	\centering
	\begin{tabular}{cc}
		\hspace{-0.4cm}
		\begin{adjustbox}{valign=t}
			\begin{tabular}{c}
				\includegraphics[width=0.151\textwidth]{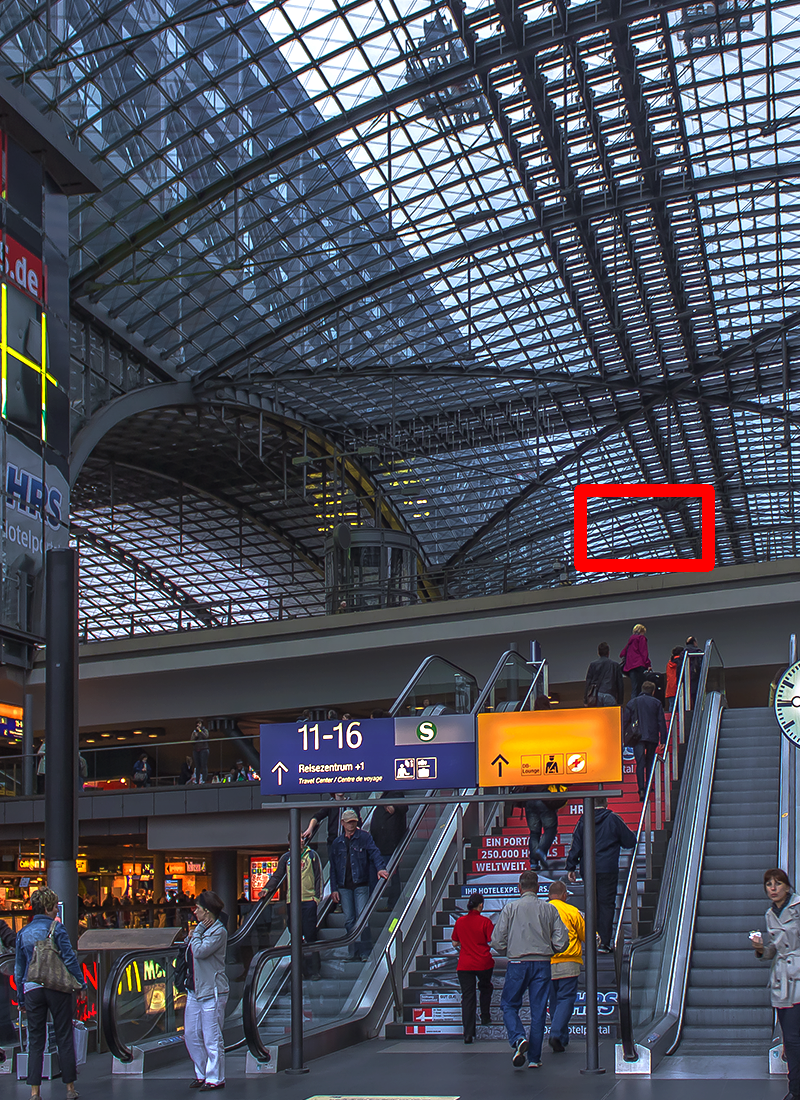}
				\\
				img\_0035 (Left)
			\end{tabular}
		\end{adjustbox}
		\hspace{-0.4cm}
		\begin{adjustbox}{valign=t}
			\begin{tabular}{cccccc}
				\includegraphics[width=0.161\textwidth]{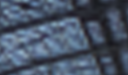} \hspace{-3mm} &
				\includegraphics[width=0.161\textwidth]{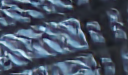} \hspace{-3mm} &
				\includegraphics[width=0.161\textwidth]{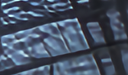} \hspace{-3mm} &
				\includegraphics[width=0.161\textwidth]{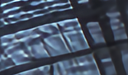} \hspace{-3mm} &
				\includegraphics[width=0.161\textwidth]{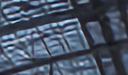} \hspace{-3mm} 
%
				
				\\
				Bicubic \hspace{-3mm} &
				StereoSR \hspace{-3mm} &
				EDSR \hspace{-3mm} &
				RCAN\hspace{-3mm} &
				SRRes+SAM \hspace{-3mm}
				\\

				\includegraphics[width=0.161\textwidth]{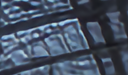} \hspace{-3mm} &
				\includegraphics[width=0.161\textwidth]{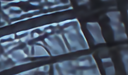} \hspace{-3mm} &
				\includegraphics[width=0.161\textwidth]{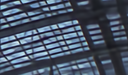} \hspace{-3mm}   &
				\includegraphics[width=0.161\textwidth]{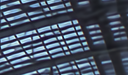} \hspace{-3mm} &
				\includegraphics[width=0.161\textwidth]{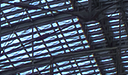} \hspace{-3mm} 
				\\ 
				iPASSR \hspace{-3mm} &
				SSRDE-FNet  \hspace{-3mm} &
				NAFSSR-L \hspace{-3mm} &
				SwinFIRSSR(ours) \hspace{-3mm} &
				Reference \hspace{-3mm}
				\\
			\end{tabular}
		\end{adjustbox}
		\vspace{1mm}
		\\
		
		\hspace{-0.4cm}
		\begin{adjustbox}{valign=t}
			\begin{tabular}{c}
				\includegraphics[width=0.151\textwidth]{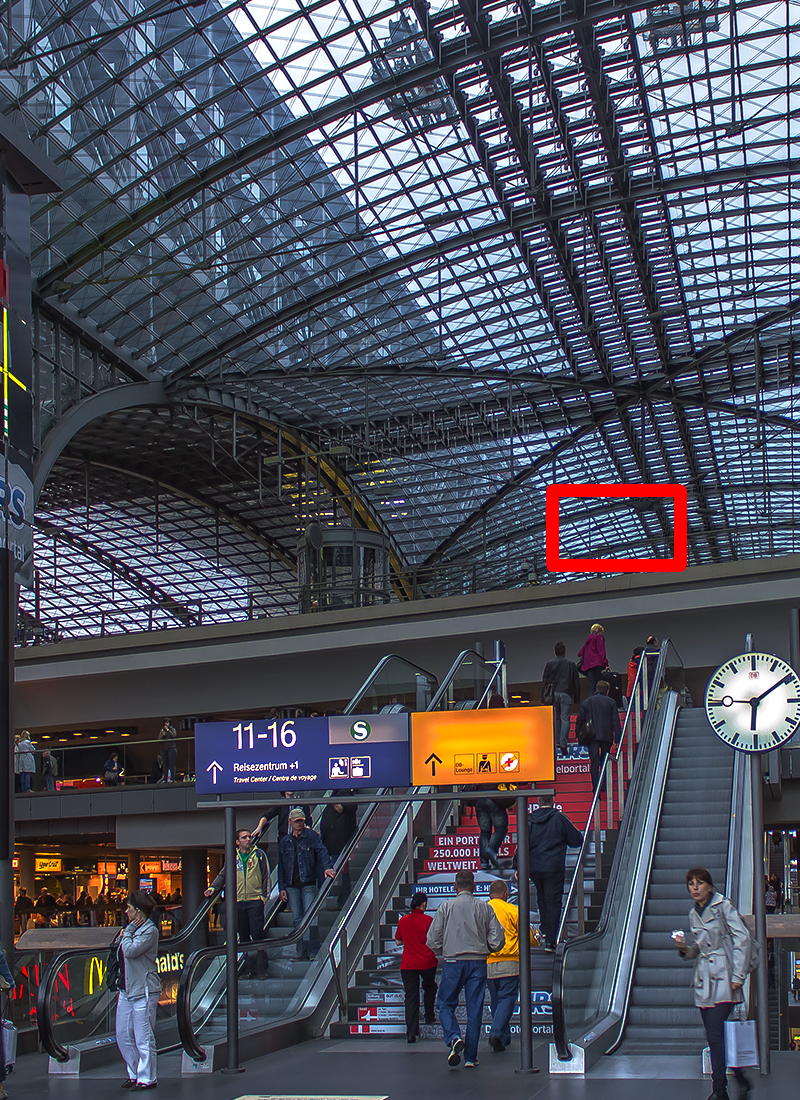}
				\\
				img\_0035 (Right)
			\end{tabular}
		\end{adjustbox}
		\hspace{-0.4cm}
		\begin{adjustbox}{valign=t}
			\begin{tabular}{cccccc}
				
				\includegraphics[width=0.161\textwidth]{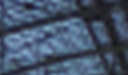} \hspace{-3mm} &
				\includegraphics[width=0.161\textwidth]{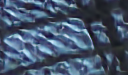} \hspace{-3mm} &
				\includegraphics[width=0.161\textwidth]{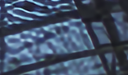} \hspace{-3mm} &
				\includegraphics[width=0.161\textwidth]{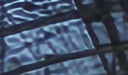} \hspace{-3mm} &
				\includegraphics[width=0.161\textwidth]{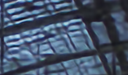} \hspace{-3mm}
				\\
				
				Bicubic \hspace{-3mm} &
				StereoSR \hspace{-3mm} &
				EDSR \hspace{-3mm} &
				RCAN \hspace{-3mm} &
				SRRes+SAM \hspace{-3mm}
				\\
				
				\includegraphics[width=0.161\textwidth]{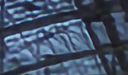} \hspace{-3mm} &
				\includegraphics[width=0.161\textwidth]{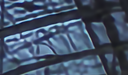} \hspace{-3mm} &
				\includegraphics[width=0.161\textwidth]{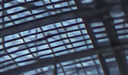} \hspace{-3mm}   &
				\includegraphics[width=0.161\textwidth]{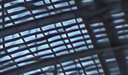} \hspace{-3mm} &
				\includegraphics[width=0.161\textwidth]{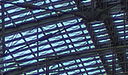} \hspace{-3mm} 
				\\ 
				
				iPASSR \hspace{-3mm} &
				SSRDE-FNet  \hspace{-3mm} &
				NAFSSR-L \hspace{-3mm} &
				SwinFIRSSR(ours) \hspace{-3mm} &
				Reference \hspace{-3mm}
				\\
			\end{tabular}
		\end{adjustbox}
		\vspace{1mm}
		\\
	\end{tabular}
	\vspace{-3mm}
	\caption{Visual results ($\times$4) achieved by different methods on the Flickr1024 dataset (\textbf{stereo image SR}). 
	}
	\label{fig:flickr1024}
	\vspace{-3mm}
\end{figure*}
\begin{table*}[!t]
	\centering
	
	\resizebox{\textwidth}{!}
	{
		\begin{tabular}{lccccccccc}
			\toprule
			\multirow{2}*{Method} & \multirow{2}*{Scale} & \multirow{2}*{$\#P$} & \multicolumn{3}{c}{\textit{Left}} & \multicolumn{4}{c}{$\left(\textit{Left}+\textit{Right}\right)/2$}\\
			\cmidrule(lr){4-6} \cmidrule(lr){7-10}
			&      &           & KITTI 2012 & KITTI 2015 & Middlebury & KITTI 2012 & KITTI 2015 & Middlebury & Flickr1024\\
			\midrule
			VDSR & $\times$2 & 0.66M & 30.17$/$0.9062 & 28.99$/$0.9038 & 32.66$/$0.9101 & 30.30$/$0.9089 & 29.78$/$0.9150& 32.77$/$0.9102 & 25.60$/$0.8534\\
			EDSR & $\times$2 & 38.6M & 30.83$/$0.9199 & 29.94$/$0.9231 & 34.84$/${0.9489} &30.96$/$0.9228 & 30.73$/$0.9335 & {34.95}$/${0.9492} & {28.66}$/$0.9087 \\
			RDN & $\times$2 & 22.0M  & 30.81$/$0.9197 & 29.91$/$0.9224 & {34.85}$/$0.9488 &30.94$/$0.9227 & 30.70$/$0.9330 & 34.94$/$0.9491 & 28.64$/$0.9084 \\
			RCAN & $\times$2 & 15.3M & 30.88$/$0.9202 & 29.97$/$0.9231 & 34.80$/$0.9482 & 31.02$/$0.9232 & 30.77$/$0.9336 & 34.90$/$0.9486 & 28.63$/$0.9082 \\
			StereoSR & $\times$2 &1.08M & 29.42$/$0.9040 & 28.53$/$0.9038 & 33.15$/$0.9343 & 29.51$/$0.9073 & 29.33$/$0.9168 & 33.23$/$0.9348 & 25.96$/$0.8599 \\
			PASSRnet & $\times$2 & 1.37M & 30.68$/$0.9159 & 29.81$/$0.9191 & 34.13$/$0.9421 & 30.81$/$0.9190 & 30.60$/$0.9300 & 34.23$/$0.9422 & 28.38$/$0.9038 \\
			 IMSSRnet & $\times$2 & 6.84M & 30.90$/$- & 29.97$/$- & 34.66$/$- & 30.92$/$- & 30.66$/$- & 34.67$/$- & -$/$- \\
			iPASSR & $\times$2 & 1.37M & {30.97}$/${0.9210} & {30.01}$/${0.9234} & 34.41$/$0.9454 & {31.11}$/${0.9240} & {30.81}$/${0.9340} & 34.51$/$0.9454 & 28.60$/${0.9097} \\
			SSRDE-FNet   & $\times$2 & 2.10M & {31.08}$/${0.9224} & {30.10}$/${0.9245} & {35.02}$/${0.9508} & {31.23}$/${0.9254} & {30.90}$/${0.9352} & {35.09}$/${0.9511} & {28.85}$/${0.9132} \\
			NAFSSR-T & $\times$2 & 0.45M  & 31.12$/$0.9224	& 30.19$/$0.9253	 & 34.93$/$0.9495 & 31.26$/$0.9254& 30.99$/$0.9355 & 35.01$/$0.9495 & 28.94$/$0.9128\\
			NAFSSR-S & $\times$2 & 1.54M  & 31.23$/$0.9236	& 30.28$/$0.9266	 & 35.23$/$0.9515 & 31.38$/$0.9266& 31.08$/$0.9367 & 35.30$/$0.9514 & 29.19$/$0.9160\\
			NAFSSR-B & $\times$2 & 6.77M  & {31.40}$/${0.9254} & {30.42}$/${0.9282} & {35.62}$/${0.9545} & {31.55}$/${0.9283} & {31.22}$/${0.9380} & {35.68}$/${0.9544} & {29.54}$/${0.9204} \\
			NAFSSR-L & $\times$2 & 23.79M  & 31.45$/${0.9261} & {30.46}$/${0.9289} & {35.83}$/${0.9559} & {31.60}$/${0.9291} & {31.25}$/${0.9386} & {35.88}$/${0.9557} & {29.68}$/${0.9221} \\
			\textbf{SwinFIRSSR} (Ours) & $\times$2 & 23.94M  & {31.65}$/${0.9293} & {30.66}$/${0.9321} & {36.48}$/${0.9601} & {31.79}$/${0.9321} & {31.45}$/${0.9413} & {36.52}$/${0.9598} & {30.14}$/${0.9286} \\
			\midrule
			\midrule
			VDSR &  $\times$4 & 0.66M & 25.54$/$0.7662 & 24.68$/$0.7456 & 27.60$/$0.7933 & 25.60$/$0.7722 & 25.32$/$0.7703 & 27.69$/$0.7941 & 22.46$/$0.6718 \\
			EDSR &  $\times$4 & 38.9M & 26.26$/$0.7954 & 25.38$/$0.7811 & 29.15$/${0.8383} & 26.35$/$0.8015 & 26.04$/$0.8039 & 29.23$/$0.8397 & 23.46$/$0.7285 \\
			RDN &  $\times$4 & 22.0M  & 26.23$/$0.7952 & 25.37$/$0.7813 & 29.15$/$0.8387 & 26.32$/$0.8014 & 26.04$/$0.8043 & 29.27$/${0.8404} & 23.47$/${0.7295} \\
			RCAN &  $\times$4 & 15.4M & 26.36$/$0.7968 & 25.53$/$0.7836 & {29.20}$/$0.8381 & 26.44$/$0.8029 & 26.22$/$0.8068 & {29.30}$/$0.8397 & {23.48}$/$0.7286 \\
			StereoSR  &  $\times$4 & 1.42M   & 24.49$/$0.7502 & 23.67$/$0.7273 &27.70$/$0.8036 & 24.53$/$0.7555 & 24.21$/$0.7511 & 27.64$/$0.8022 & 21.70$/$0.6460 \\
			PASSRnet  &  $\times$4 & 1.42M   & 26.26$/$0.7919 & 25.41$/$0.7772 &28.61$/$0.8232 & 26.34$/$0.7981 & 26.08$/$0.8002 & 28.72$/$0.8236 & 23.31$/$0.7195 \\
			SRRes+SAM  &  $\times$4 & 1.73M  & 26.35$/$0.7957 & 25.55$/$0.7825 & 28.76$/$0.8287 & 26.44$/$0.8018 & 26.22$/$0.8054 & 28.83$/$0.8290 & 23.27$/$0.7233 \\
			IMSSRnet &  $\times$4 & 6.89M  & 26.44$/$- & 25.59$/$- & 29.02$/$- & 26.43$/$- & 26.20$/$- & 29.02$/$- & -$/$- \\
			iPASSR  &  $\times$4 & 1.42M  & {26.47}$/${0.7993} & {25.61}$/${0.7850} & 29.07$/$0.8363 & {26.56}$/${0.8053} & {26.32}$/${0.8084} & 29.16$/$0.8367 & 23.44$/$0.7287 \\
			SSRDE-FNet  & $\times$4 & 2.24M  & {26.61}$/${0.8028} & {25.74}$/${0.7884} & {29.29}$/${0.8407} & {26.70}$/${0.8082} & {26.43}$/${0.8118} & {29.38}$/${0.8411} & {23.59}$/${0.7352} \\
			NAFSSR-T & $\times$4 & 0.46M  & 26.69$/$0.8045	& 25.90$/$0.7930	 & 29.22$/$0.8403 & 26.79$/$0.8105& 26.62$/$0.8159 & 29.32$/$0.8409 & 23.69$/$0.7384\\
			NAFSSR-S & $\times$4 & 1.56M  & {26.84}$/${0.8086}	& {26.03}$/${0.7978}	 & 29.62$/$0.8482 & {26.93}$/$0.8145& {26.76}$/$0.8203 & 29.72$/$0.8490 & 23.88$/$0.7468\\
			NAFSSR-B  & $\times$4 & 6.80M  & {26.99}$/${0.8121} & {26.17}$/${0.8020} & {29.94}$/${0.8561} & {27.08}$/${0.8181} & {26.91}$/${0.8245} & {30.04}$/${0.8568} & {24.07}$/${0.7551} \\
			NAFSSR-L & $\times$4 & 23.83M  & 27.04$/${0.8135} & {26.22}$/${0.8034} & {30.11}$/${0.8601} & {27.12}$/${0.8194} & {26.96}$/${0.8257} & {30.20}$/${0.8605} & {24.17}$/${0.7589} \\
			\textbf{SwinFIRSSR} (Ours) & $\times$4 & 24.09M  & {27.06}$/${0.8175} & {26.15}$/${0.8062} & {30.33}$/${0.8676} & {27.16}$/${0.8235} & {26.89}$/${0.8283} & {30.44}$/${0.8687} & {24.29}$/${0.7681} \\
			\bottomrule
	\end{tabular}}
 \caption{Quantitative results achieved by different methods on the KITTI 2012~\cite{geiger2012we}, KITTI 2015~\cite{menze2015object}, Middlebury~\cite{scharstein2014high}, and Flickr1024~\cite{wang2019learning} datasets on the RGB space for \textbf{stereo image SR}. $\#P$ represents the number of parameters of the networks. 
		Here, PSNR$/$SSIM values achieved on both the left images (i.e., \textit{Left}) and a pair of stereo images (i.e., $\left(\textit{Left}+\textit{Right}\right)/2$) are reported. } 
	\label{tab:SwinFIRSSR_stereosr_quantitative_results}
\end{table*}

\subsubsection{Stereo Image Super-Resolution}

We conduct a series of experiments in the stereo image SR and compare SwinFIRSSR to other SR methods. Single image SR methods include EDSR, RCAN, and RDN~\cite{2018Residual}, whereas stereo image SR methods include StereoSR~\cite{2018Enhancing}, PASSRnet~\cite{2019Learning}, SRRes+SAM~\cite{2020A}, IMSSRnet~\cite{lei2020deep}, iPASSR~\cite{2020Deep}, SSRDE-FNet~\cite{2021Feedback} and NAFSSR~\cite{chu2022nafssr}. 
All models are trained on 800 Flickr1024 and 60 Middlebury images. 
Our SwinFIRSSR surpasses all single and stereo image SR methods, as shown in Table~\ref{tab:SwinFIRSSR_stereosr_quantitative_results}. When the number of parameters of our method is comparable to the SOTA method NAFSSR-L, our SwinFIRSSR performs better for $\times$2 stereo SR than NAFSSR-L by 0.19 dB, 0.20 dB, 0.64 dB and 0.46 dB on the KITTI2012~\cite{geiger2012we}, KITTI2015~\cite{menze2015object}, Middlebury~\cite{scharstein2014high} and Flickr1024 testing dataset, respectively. 
The visual comparison results are shown in Figure~\ref{fig:flickr1024}. The SR images reconstructed by our SwinFIRSSR are clearer and offer a wealth of details and textures. 
The quantitative and qualitative results on stereo SR datasets demonstrate the effectiveness and robustness of our SwinFIRSSR.


\subsection{Ablation Study}
\subsubsection{Impact of RFB}
\label{sec:RFB}
We try using the original FB, named Spectral Transform in~\cite{suvorov2022resolution}, to replace the convolution module in the RSTB to extract global features from the low resolution image. However, this does not lead to the expected performance improvement for image SR, as shown in the Table~\ref{tab:SFB_ABS}. It can be seen intuitively from the LAM that local information has a higher contribution to image reconstruction. Therefore, we replace FB with SFB, that is, only adding a simple residual module to extract the local features of the input image can significantly improve the performance. At the same time, we also explore the influence of SFB position in RSTB on image reconstruction. We replace the attention module in STL with SFB and hybrid SFB to obtain SFTL and HSTL, as shown in the Figure~\ref{fig:SFB_Block}. Although the performance of HSTL (37.20M) is better than that of SFB on Manga dataset, its parameters are 2.6 times that of SFB(14.03M), so we abandon this scheme.

\begin{table}[!t]
	\centering
		\scalebox{0.88}
		{
		\begin{tabular}{ccccccc}
			\hline
			& Method & Set5 & Set14 & BSD & Urban & Manga
			\\ 
			\hline
			\hline
			\multirow{3}{*}{\tabincell{c}{Model \\ Design}}
			& SwinIR
			& 32.72
			& 28.94
			& 27.83
			& 27.07
			& 31.67
			\\
			& FB
			& 32.65
			& 28.98
			& 27.81
			& 26.86
			& 31.54
			\\
			& \textbf{SFB}
			& 32.78
			& 29.01
			& 27.86
			& 27.11
			& 31.69
			\\
			\hdashline
			\multirow{2}{*}{Position}
			& SFTL
			& 32.61
			& 28.89
			& 27.78
			& 26.91
			& 31.49
			\\
			& HSTL
			& 32.75
			& 28.98
			& 27.85
			& 27.07
			& 31.76
			\\
			\hline             
		\end{tabular}
	}
		\caption{Impact of SFB with different model design and position.}
		\label{tab:SFB_ABS}
\end{table}

\subsubsection{Impact of Data Augmentations}
Radu~\etal propose seven ways of data augmentations methods without altered content. They contend that other data augmentations will introduce new pixels and degrade super-resolution performance. In addition to flip and rotation, we revisit the effect of data augmentations on image super-resolution. The channel shuffle and Mixup data augmentations improve the PSNR from 32.78 dB to 32.93 dB, which is 0.15 dB better than flip and rotation, as shown in Figure~\ref{fig:roadmap}. The experiments demonstrate that the data augmentations methods based on inserting new pixels sometimes improve the performance of SR, breaking people's previous cognition. However, not all data augmentations can improve image super-resolution performance. For example, CutMix and CutMixup boost the performance of image classification, but they have a drawback in the low-level tasks: they obliterate visual continuity. This drawback makes the image lose semantic information and reduces the available and useful information. 

\begin{figure}[h]
	\centering
	\includegraphics[width=0.78\linewidth]{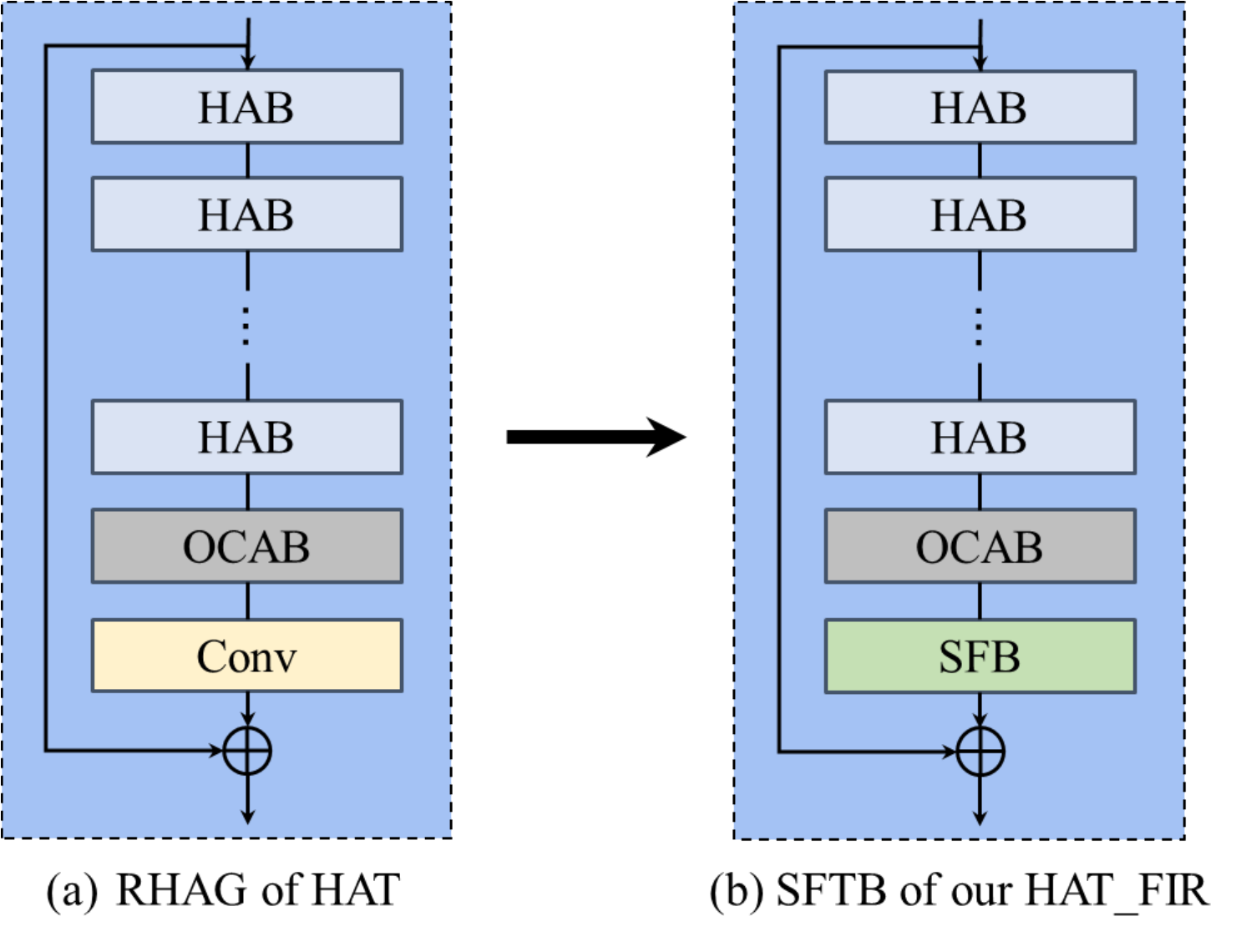}
	\caption{(a) Residual Hybrid Attention Group (RHAG) in the HAT. (b) Swin Fourier Transformer Block (SFTB) in our HAT\_FIR. We replace the convolution (3$\times$3) in RHAG of HAT with SFB to get our SFTB.} 
	\label{fig:SFTB}
\end{figure}

\begin{table}[!t]
	\centering
		\scalebox{0.81}
		{
		\begin{tabular}{lcccccc}
			\hline
			Method & S & Set5 & Set14 & BSD & Urban & Manga
			\\ 
			\hline
			\hline
			HAT & $\times$2
			& 38.63
			& 34.86
			& 32.62
			& 34.45
			& 40.26
			\\
			\textbf{HAT\_FIR} (Ours) & $\times$2
			& 38.63
			& 34.63
			& 32.62
			& 34.47
			& 40.39
			\\
			\textbf{+FE} (Ours) & $\times$2
			& 38.63
			& 34.70
			& 32.63
			& 34.55
			& 40.41
			\\
			\hdashline
			HAT$^\dagger$ & $\times$2
			& 38.73
			& 35.13
			& 32.69
			& 34.81
			& 40.71
			\\
			\textbf{HAT\_FIR}$^\dagger$ (Ours) & $\times$2
			& 38.74
			& 35.16
			& 32.71
			& 34.92
			& 40.72
			\\
			\textbf{+FE} (Ours) & $\times$2
			& 38.74
			& 35.17
			& 32.71
			& 34.94
			& 40.77
			\\
			\hline
			\hline
			HAT & $\times$3
			& 35.06
			& 31.08
			& 29.54
			& 30.23
			& 35.53
			\\
			\textbf{HAT\_FIR} (Ours) & $\times$3
			& 35.13
			& 31.15
			& 29.54
			& 30.40
			& 35.60
			\\
			\textbf{+FE} (Ours) & $\times$3
			& 35.13
			& 31.17
			& 29.55
			& 30.42
			& 35.62
			\\
			\hdashline
			HAT$^\dagger$ & $\times$3
			& 35.16
			& 31.33
			& 29.59
			& 30.70
			& 35.84
			\\
			\textbf{HAT\_FIR}$^\dagger$ (Ours) & $\times$3
			& 35.20
			& 31.36
			& 29.60
			& 30.74
			& 35.90
			\\
			\textbf{+FE} (Ours) & $\times$3
			& 35.21
			& 31.37
			& 29.60
			& 30.77
			& 35.92
			\\
			\hline
			\hline
			HAT & $\times$4
			& 33.04
			& 29.23
			& 28.00
			& 27.97
			& 32.48
			\\
			\textbf{HAT\_FIR} (Ours) & $\times$4
			& 33.09
			& 29.23
			& 28.01
			& 28.13
			& 32.64
			\\
			\textbf{+FE} (Ours) & $\times$4
			& 33.15
			& 29.23
			& 28.01
			& 28.14
			& 32.68
			\\
			\hdashline
			HAT$^\dagger$ & $\times$4
			& 33.18
			& 29.38
			& 28.05
			& 28.37
			& 32.87
			\\
			\textbf{HAT\_FIR}$^\dagger$ (Ours) & $\times$4
			& 33.26
			& 29.44
			& 28.07
			& 28.40
			& 33.02
			\\
			\textbf{+FE} (Ours) & $\times$4
			& 33.29
			& 29.44
			& 28.07
			& 28.43
			& 33.03
			\\
			\hline             
		\end{tabular}}
		\caption{Quantitative comparison with HAT.}
		\label{tab:HATFIR_quantitative_results}
\end{table}

\begin{table}[!t]
	\centering
	\scalebox{0.78}
	{
		\begin{tabular}{cccccc}
			\hline
			Task & Method & S & DIV2K (val) & DIV2K (test) & 
			\\ 
			\hline
			\multirow{2}{*}{SR}
			& ABPN & $\times$3
			& 30.14
			& 29.87
			& 
			\\
			& \textbf{ABPN\_FIR} & $\times$3
			& 30.18
			& 29.95
			& 
			\\
			\hline
			\hline
			& Method & $\sigma$ & Set12 & BSD68 & Urban100
			\\ 
			\hline
			\multirow{2}{*}{DN-G}
			& SwinIR & 15
			& 33.36
			& 31.97
			& 33.70
			\\
			& \textbf{SwinFIR} & 15
			& 33.42
			& 32.00
			& 33.85
			\\
			\hline
			\hline
			& Method & $\sigma$ & CBSD68 & McMaster & Urban100
			\\ 
			\hline
			\multirow{2}{*}{DN-C}
			& SwinIR & 15
			& 34.42
			& 35.61
			& 35.13
			\\
			& \textbf{SwinFIR} & 15
			& 34.43
			& 35.64
			& 35.34
			\\
			\hline     
			\hline
			& Method & $q$ & Classic5 &  & 
			\\ 
			\hline
			\multirow{2}{*}{JPEG}
			& SwinIR & 10
			& 30.27
			& 
			& 
			\\
			& \textbf{SwinFIR} & 10
			& 30.65
			& 
			& 
			\\
			\hline          
	\end{tabular}}
	\caption{Quantitative comparison on different low-level tasks.}
	\label{tab:ABPN}
\end{table}

\subsubsection{Robustness}
HAT~\cite{chen2022activating} propose a Hybrid Attention Transformer based on SwinIR and achieve the SOTA performance in image super-resolution task. So we also put HAT in our roadmap to verify the adaptability and robustness of our methods, as shown in Figure~\ref{fig:SFTB} and Table~\ref{tab:HATFIR_quantitative_results}. Our training set and configuration are consistent with HAT in order to conduct a fair comparison.
Our HAT\_FIR improve the PSNR from 32.48 dB to 32.64 dB on Manga109 dataset ($\times4$), which is 0.16 dB better than HAT. If we add the feature ensemble strategy into HAT\_FIR, the performance will be further improved. 

ABPN~\cite{du2021anchor} is a simple but efficient image super-resolution method and wins the first place on Mobile AI 2021 Real-Time Image Super-Resolution Challenge (CVPR2021). We perform our method on ABPN without SFB to verify the robustness in the non-Transformer structure, as shown in Table~\ref{tab:ABPN}. We improve the PSNR by 0.08 dB on DIV2K dataset without spending any additional time on inference. We also perform our SwinFIR on grayscale image denoising (DN-G), color image denoising (DN-C) and JPEG compression artifact reduction (JPEG) benchmark datasets. All experimental results demonstrate that our method can improve the image restoration performance on low-level tasks.

\subsubsection{Impact of Feature Ensemble}
SwinIR has indicated that post-processing methods such as self-ensemble can considerably improve the performance of image super-resolution. However, self-ensemble has a substantial disadvantage in that it takes longer to do inference. 
It is very unfriendly for the on-device side or time-sensitive scenarios. 
Inspired by self-ensemble, which ensembles the results of image super-resolution, we ensemble the parameters of the trained models , named feature ensemble. 
Experimental results demonstrate that our feature ensemble steadily improves the super-resolution performance with only once inference, as shown in the Table~\ref{tab:HATFIR_quantitative_results}.

\section{Conclusion}
In this paper, we revisit how to improve the performance of image restoration. Firstly, we revisit the limitations of long-dependent modeling capabilities of SwinIR and propose an efficient global feature extractor based on Fast Fourier Convolution (FFC), named SwinFIR. Furthermore, we also revisit other strategies for improving SR performance, including data augmentation, loss function, pre-training, and feature ensemble. The results demonstrate that our SwinFIR has a wider receptive field than SwinIR, and takes advantage of long dependencies based on FFC to use more pixels for better performance. In particular, our feature ensemble strategy steadily improves performance without lengthening the training and testing periods. Extensive experiments on popular benchmarks demonstrate that our SwinFIR surpasses current models and our methods can significantly improve the performance of image restoration.

{\small
\bibliographystyle{ieee_fullname}
\bibliography{egbib}
}

\end{document}